\documentclass[final,5p,times,twocolumn]{elsarticle}

\usepackage{hyperref}
\usepackage{subfig}
\usepackage{multirow}
\usepackage[fleqn]{amsmath}
\usepackage{booktabs}
\usepackage{algorithmic}
\usepackage{enumerate}
\usepackage{array}
\usepackage{gensymb}
\usepackage{stfloats}
\fnbelowfloat
\usepackage{url}
\usepackage{color}
\usepackage{stfloats}
\usepackage{amssymb}
\usepackage{arydshln}

\usepackage{booktabs}
\usepackage{array, caption, threeparttable}
\usepackage[font=small,labelfont=bf,labelsep=none]{caption}
\usepackage[marginal]{footmisc}

\usepackage{enumitem}
\usepackage{siunitx}

\usepackage{lineno}

\usepackage{listings}
\usepackage{xcolor}
\usepackage{float}

\usepackage[ruled,vlined]{algorithm2e}

\usepackage{colortbl}
\usepackage{graphicx}
\newcommand{\filldash}[1]{#1~\xleaders\hbox{-}\hfill\kern0pt}

\begin{document}

	\begin{frontmatter}

        \title{WHU-Infra3D: A Full-stack Multi-modal Dataset and Benchmark \\ for 3D Roadside Infrastructure Inventory}
        
		\author[liesmars]{Chong Liu\footnotemark[1]}

        \author[liesmars]{Luxuan Fu\footnotemark[1]}

        \author[liesmars]{Xuyu Feng}
        
        \author[liesmars]{Zhen Dong\corref{correspondingauthor}}
        
        \author[liesmars]{Bisheng Yang}
		
		\cortext[correspondingauthor]{Corresponding authors.}
		\address[liesmars]{State Key Laboratory of Information Engineering in Surveying, Mapping and Remote Sensing (LIESMARS), Wuhan University, Wuhan 430079, China}
	
        \begin{abstract}
            The paradigm of digital twin cities is shifting from coarse visual mapping toward more precise and actionable digitization of urban assets. However, existing datasets predominantly focus on coarse visual perception, lacking the strict multi-modal alignment and attribute and status diagnosis required for automated infrastructure maintenance. To bridge this gap, we introduce \textbf{WHU-Infra3D}, a large-scale, multi-modal benchmark dataset dedicated to roadside infrastructure inventory. Covering 53.8 km across three cities, WHU-Infra3D uniquely integrates panoramic imagery and LiDAR point clouds with rigorous 2D-3D instance association and cross-frame tracking. Comprising over 175k multi-view 2D bounding boxes alongside thousands of 3D infrastructure instances, the dataset provides over 181k detailed attribute and status annotations (e.g., rust, occlusion) to empower operational health assessment. We establish comprehensive baselines across five core tasks: 2D detection, 2D cross-view matching, 3D geo-identification, 3D point cloud segmentation, and attribute recognition. Extensive evaluations expose significant cross-city domain gaps and inherent vulnerabilities of current models on long-tailed defective statuses, establishing WHU-Infra3D as an essential testbed for advancing scalable, AI-driven urban infrastructure inventory and lifecycle management. The WHU-Infra3D dataset is available at \url{https://github.com/WHU-USI3DV/WHU-Infra3D}.
		\end{abstract}

		\begin{keyword}
        Roadside infrastructure \sep Multi-modal dataset \sep 3D inventory \sep Attribute recognition
		\end{keyword}

	\end{frontmatter}

    \footnotetext[1]{Equally contributed.}
	
    \section{Introduction}\label{sec:intro}
    
\begin{figure}[!htbp]
            \centering{
            \includegraphics[width=\columnwidth]{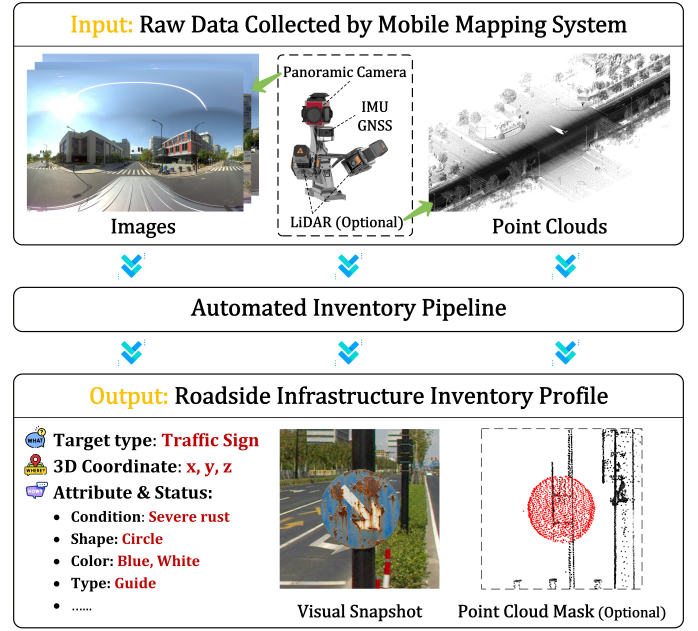}}%
            \caption{\textbf{Conceptual illustration of 3D Roadside Infrastructure Inventory.} It transforms unstructured multi-modal sensory inputs (panoramic images and LiDAR point clouds) into a structured digital asset database. The output inventory profile provides a holistic representation for each asset, encompassing its visual snapshot, 3D point cloud mask, global coordinate, and fine-grained attributes and statuses.}
            \label{fig:inv_def}
        \end{figure}

        Building high-fidelity Digital Twin Cities increasingly demands moving beyond coarse 3D geometric reconstruction toward the precise semantic and operational digitization of physical urban assets. As critical urban elements, roadside infrastructure---ranging from street lights and traffic signs to manhole covers and fire hydrants---plays a pivotal role in ensuring public safety and municipal efficiency. Effectively managing these facilities relies on a comprehensive \textbf{infrastructure inventory}. As formally defined in Fig.~\ref{fig:inv_def}, we conceptualize this inventory as an automated pipeline that transforms unstructured multi-modal sensory data---routinely acquired by Mobile Mapping Systems (MMS) equipped with LiDAR and panoramic cameras---into a highly structured digital asset profile. Achieving this objective requires more than merely discovering basic semantic categories and 3D spatial coordinates; it crucially demands a deeper cognitive diagnosis of each asset's \textbf{attributes} (e.g., shape, material) and \textbf{statuses} (e.g., damaged, tilted, or occluded). Although modern MMS hardware successfully guarantees high-precision data capture, the critical lack of comprehensive, multi-modal datasets capable of supervising the full progression from raw ``visual perception'' to fine-grained ``cognitive diagnosis'' remains a primary bottleneck, hindering the transition from manual inspection to AI-driven intelligent maintenance.

        \begin{figure*}[t]
            \centering{
            \includegraphics[width=\textwidth]{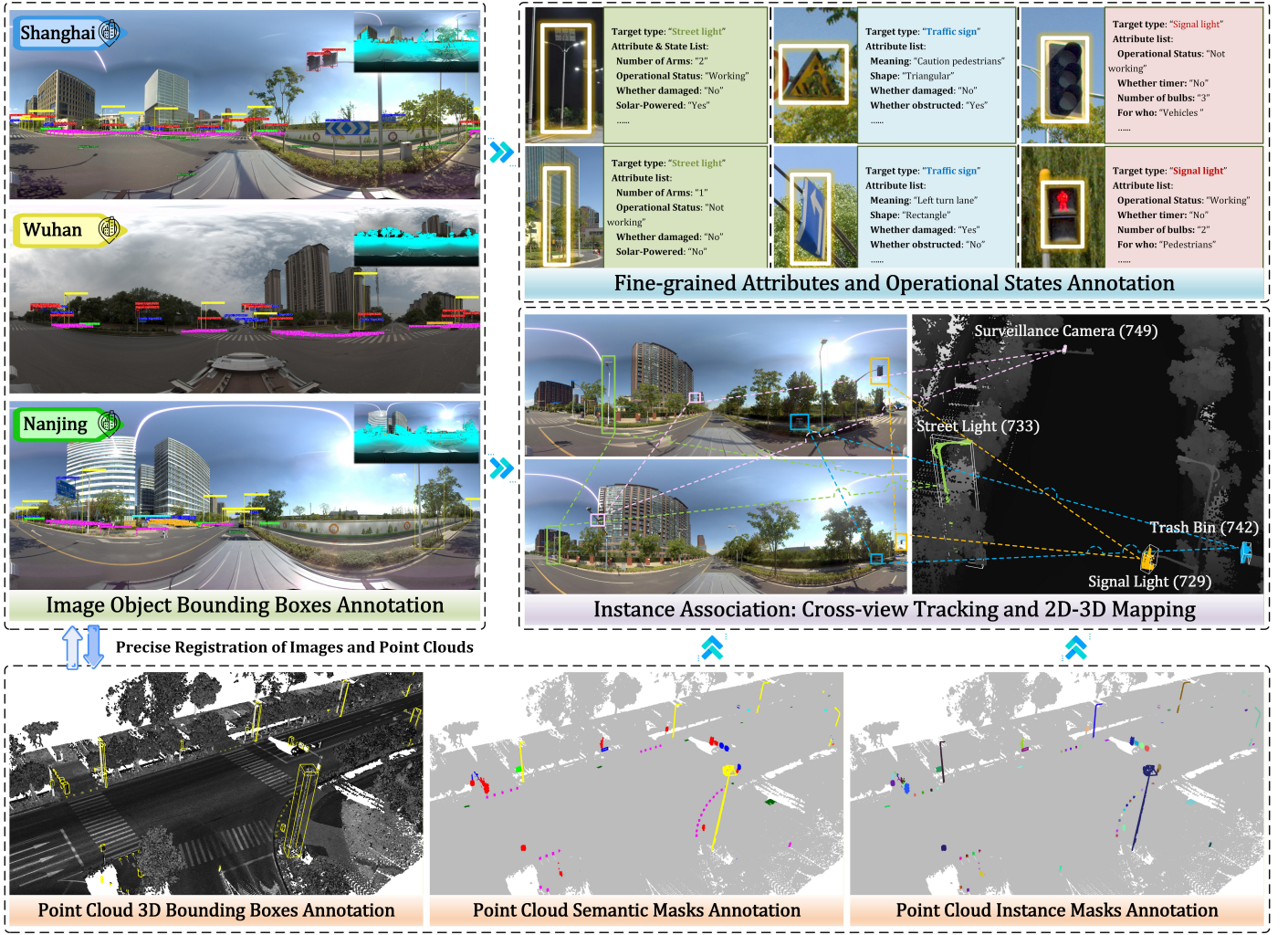}}%
            \caption{\textbf{Overview of the WHU-Infra3D dataset.} The dataset provides a comprehensive platform for urban infrastructure inventory by integrating (Left) cross-city panoramic imagery with 2D bounding box annotations and (Bottom) LiDAR point clouds with rich 3D annotations (bounding boxes, semantic masks, and instance masks). (Center) The core feature is \textbf{Instance Association}, which establishes consistent cross-frame tracking IDs across sparse views and precise 2D-3D mapping between image objects and point cloud instances. (Right) Beyond geometry, the dataset offers attribute and status annotations (e.g., identifying damaged or occluded facilities), enabling intelligent asset health assessment. WHU-Infra3D bridges the gap between raw multi-modal perception and actionable infrastructure management.}
            \label{fig:teaser}
        \end{figure*}

        Existing datasets, despite their contributions, fail to fully address the requirements of fine-grained infrastructure management. First, autonomous driving datasets like NuScenes \citep{caesar2020nuscenes} prioritize dynamic traffic participants (e.g., vehicles, pedestrians), often treating diverse static infrastructure as background or grouping them into coarse categories. Second, urban scene understanding datasets aim to parse the entire environment via semantic segmentation but suffer from significant task and modality gaps. Their primary goal is pixel-level or point-level classification rather than object-level asset management. Furthermore, 2D datasets (e.g., Cityscapes \citep{cordts2016cityscapes}) lack the 3D spatial information required for mapping, while 3D datasets (e.g., Semantic3D \citep{hackel2017semantic3d}) often provide only point clouds without corresponding imagery or image-level annotations, hindering the utilization of rich visual textures for recognition. Third, asset perception datasets focus on specific facilities but exhibit critical deficiencies for systematic inventory. Datasets like TT100K \citep{zhu2016traffic} are limited to single-frame detection and lack cross-frame tracking IDs, making it impossible to perform asset counting or deduplication across image sequences. While some recent datasets like ARTSv2 \citep{wilson2022object} provide tracking IDs and 2D geo-coordinates, they still lack precise 3D localization and, most critically, fail to capture \textbf{status} (e.g., rust, breakage, occlusion). In summary, there is no existing benchmark that simultaneously offers multi-modal data, precise 3D localization, and comprehensive attribute and status descriptions for generic roadside infrastructure.
        
        To bridge this gap, we present \textbf{WHU-Infra3D}. As illustrated in Fig. \ref{fig:teaser}, WHU-Infra3D dataset establishes a \textbf{holistic paradigm for intelligent infrastructure inventory}. First, it provides a \textbf{multi-modal and multi-task data foundation}: spanning a 53.8 km trajectory across three diverse Chinese cities (Wuhan, Shanghai, and Nanjing), it integrates high-resolution panoramic imagery and high-density LiDAR point clouds collected by a high-end MMS under high-precision registration. Second, a \textbf{full-stack fine-grained annotation system} is introduced, providing precise mappings between 2D bounding boxes and 3D point cloud instances, along with cross-frame tracking IDs. Crucially, a detailed attribute schema is defined covering both attributes and statuses (e.g., identifying if a bollard is broken), directly empowering algorithms to perform status assessments on city assets. Finally, the dataset reflects \textbf{real-world complexity}, preserving natural long-tail distributions of defective cases and capturing significant cross-city domain gaps in infrastructure design, posing new challenges for algorithmic generalization.
        
        The main contributions of this paper are summarized as follows:
        \begin{itemize}
            \item \textbf{Comprehensive Multi-Modal Dataset.} We introduce WHU-Infra3D, a pioneering multi-modal and multi-task dataset for urban infrastructure inventory. Covering 53.8 km of urban trajectories, it strictly aligns high-resolution panoramic imagery with dense LiDAR point clouds. It uniquely bridges geometric perception and intelligent diagnosis by providing precise 2D-3D mapping, cross-frame tracking, and status annotations.
            \item \textbf{Extensive Baseline Evaluations.} We formalize a rigorous five-stage benchmark: 2D detection, 2D cross-view matching, 3D geo-identification, 3D point cloud segmentation, and multi-modal attribute recognition. We extensively evaluate state-of-the-art algorithms, including Vision-Language Models (VLMs), deeply probing their geometric localization and cognitive reasoning capabilities.
            \item \textbf{Crucial Insights and Future Directions.} Our evaluations expose profound vulnerabilities in current architectures against cross-city domain shifts and long-tailed defective statuses. These insights outline promising research directions, establishing WHU-Infra3D as an indispensable testbed for advancing AI-driven lifecycle management of urban assets.
        \end{itemize}

        \begin{table*}[t]
            \centering
            \caption{Comparison of WHU-Infra3D with existing representative datasets. In the \textbf{\#Cls} column, \textbf{A(B)} denotes \textbf{A} infrastructure categories out of \textbf{B} total categories. (\textbf{Img}: Image, \textbf{PC}: Point Cloud; \textbf{Box}: Bounding Box, \textbf{Mask}: Instance Mask, \textbf{Pt}: Center Point Coordinates).}
            \label{tab:comparison}
            \resizebox{\textwidth}{!}{%
            \begin{tabular}{lc cccc cccccc}
            \toprule
            \multirow{2}{*}{\textbf{Dataset}} & \multirow{2}{*}{\textbf{Year}} & \multicolumn{4}{c}{\textbf{Data}} & \multicolumn{6}{c}{\textbf{Annotation}} \\ 
            \cmidrule(lr){3-6} \cmidrule(lr){7-12}
             & & \textbf{Scale} & \textbf{Frames} & \textbf{Img} & \textbf{PC} & \textbf{Target Focus} & \textbf{\#Cls(Inf/All)} & \textbf{2D} & \textbf{3D} & \textbf{Track ID} & \textbf{Attr.} \\ \midrule
            
            \multicolumn{12}{l}{\textit{\filldash{Autonomous Driving Datasets}}} \\
            KITTI\citep{geiger2012we} & 2012 & 39.2 km & 15k & \checkmark & \checkmark & Traffic Participants & 0(8) & Box & Box & \checkmark & - \\
            Waymo Open\citep{sun2020scalability} & 2020 & 76 km² & 230k & \checkmark & \checkmark & Traffic Participants & 1(4) & Box & Box & \checkmark & - \\
            NuScenes\citep{caesar2020nuscenes} & 2020 & - & 40k & \checkmark & \checkmark & Traffic Participants & 1(23) & Box & Box & \checkmark & - \\
            KITTI-360\citep{liao2022kitti} & 2022 & 73.7 km & 80k & \checkmark & \checkmark & Traffic Participants & 8(37) & Mask & Box,Mask & \checkmark & - \\
            Argoverse 2\citep{wilson2023argoverse} & 2023 & 200 km & 150k & \checkmark & \checkmark & Traffic Participants & 5(30) & - & Box & \checkmark & - \\ 
            \midrule
            
            \multicolumn{12}{l}{\textit{\filldash{Road Scene Understanding Datasets}}} \\
            iQmulus\citep{vallet2015terramobilita} & 2015 & 10.0 km & - & - & \checkmark & Road Scene & 1(9) & - & Mask & - & - \\
            Cityscapes\citep{cordts2016cityscapes} & 2016 & - & 25k & \checkmark & - & Urban Scene & 3(30) & Mask & - & - & - \\
            Semantic3D\citep{hackel2017semantic3d} & 2017 & 30 Scans & - & - & \checkmark & Road Scene & 0(8) & - & Mask & - & - \\
            Mapillary Vistas\citep{neuhold2017mapillary} & 2017 & - & 25k & \checkmark & - & Road Scene & 18(66) & Mask & - & - & - \\
            Paris-Lille-3D\citep{roynard2018paris} & 2018 & 2.0 km & - & - & \checkmark & Road Scene & 13(50) & - & Mask & - & - \\
            SemanticKITTI\citep{behley2019semantickitti} & 2019 & 39.2 km & 43k & \checkmark & \checkmark & Road Scene & 1(28) & - & Mask & - & - \\
            Toronto-3D\citep{tan2020toronto} & 2020 & 1.0 km & - & - & \checkmark & Road Scene & 0(8) & - & Mask & - & - \\
            WHU-Urban3D\citep{han2024whu} & 2024 & 16.0 km & - & \checkmark & \checkmark & Road Scene & 8(19) & - & Mask & - & - \\ 
            \midrule
            
            \multicolumn{12}{l}{\textit{\filldash{Asset Perception Datasets}}} \\
            GTSDB\citep{houben2013detection} & 2013 & - & 0.9k & \checkmark & - & Traffic Signs & 1(43) & Box & - & - & - \\
            TT100K\citep{zhu2016traffic} & 2016 & - & 100k & \checkmark & - & Traffic Signs & 1(221) & Box,Mask & - & - & - \\
            ARTS\citep{almutairy2019arts} & 2019 & - & 51k & \checkmark & - & Traffic Signs & 1(295) & Box & - & - & - \\
            TLG\citep{chaabane2021end} & 2021 & - & 16k & \checkmark & - & Traffic Lights & 1(1) & Box & Pt & \checkmark & - \\
            ARTSv2\citep{wilson2022object} & 2022 & - & 25k & \checkmark & - & Traffic Signs & 1(199) & Box & Pt & \checkmark & - \\
            \midrule
            
            \rowcolor{gray!15} \textbf{WHU-Infra3D} & \textbf{2025} & \textbf{53.8 km} & \textbf{5k} & \textbf{\checkmark} & \textbf{\checkmark} & \textbf{Infrastructure} & \textbf{10(10)} & \textbf{Box} & \textbf{Pt,Box,Mask} & \textbf{\checkmark} & \textbf{\checkmark} \\
            \bottomrule
            \end{tabular}%
            }
        \end{table*}
    
    \section{Related Work}
    \label{sec:related_work}
    This section reviews existing works pertinent to urban infrastructure inventory from two perspectives: datasets and methodologies. First, we categorize existing datasets into autonomous driving, urban scene understanding, and asset perception, analyzing their limitations in supporting fine-grained inventory tasks. Second, we survey state-of-the-art methods across the full pipeline---including detection, matching, localization, and attribute recognition---to establish the technical context for our proposed benchmarks.
    
    \subsection{Previous Datasets}
        With the rapid advancement of deep learning in urban digitization, a multitude of public datasets have been released. To systematically analyze their applicability to fine-grained infrastructure management, we categorize existing datasets into three groups based on their primary research focus, as summarized in Table~\ref{tab:comparison}: (1) Autonomous Driving Datasets, driven by the need for safe navigation in dynamic traffic; (2) Road Scene Understanding Datasets, aiming for holistic geometric and semantic parsing of the environment; and (3) Asset Perception Datasets, dedicated to the detection and recognition of specific road assets. While these datasets have propelled progress in their respective fields, they exhibit distinct limitations when applied to the full-stack inventory of urban infrastructure.
        
        \textbf{Autonomous Driving Datasets.} Represented by benchmarks such as KITTI \citep{geiger2012we}, Waymo Open \citep{sun2020scalability}, NuScenes \citep{caesar2020nuscenes}, and KITTI-360 \citep{liao2022kitti}, these datasets are characterized by their massive scale and rich multi-sensor data. However, their primary objective is collision avoidance, which inherently prioritizes dynamic traffic participants (e.g., vehicles, pedestrians). As shown in Table~\ref{tab:comparison}, the definition of static infrastructure in these datasets is often minimal. For instance, in NuScenes, despite having 23 categories, only traffic cone represents a distinct roadside infrastructure class of interest, while the vast majority of labels are dedicated to vehicles and humans. Even in KITTI-360, although more static categories are included, the primary focus remains on scene understanding for autonomous navigation rather than asset inventory.
        
        \textbf{Road Scene Understanding Datasets.} Datasets such as Cityscapes \citep{cordts2016cityscapes}, Semantic3D \citep{hackel2017semantic3d}, and WHU-Urban3D \citep{han2024whu} aim to parse the entire environment via semantic segmentation. While achieving comprehensive coverage, they suffer from significant task and modality gaps. Their primary goal is pixel-level or point-level classification rather than object-level asset management. Furthermore, as indicated in Table~\ref{tab:comparison}, there is a clear disconnect in multi-modal annotations: 2D datasets (e.g., Cityscapes) lack the 3D spatial information required for mapping, while 3D datasets (e.g., Semantic3D \citep{hackel2017semantic3d}) often provide only point clouds without corresponding imagery or image-level annotations, hindering the utilization of rich visual textures for recognition. Notably, none of the existing road scene understanding datasets provide high-quality annotations for both 2D imagery and 3D point clouds simultaneously, creating a barrier for effective multi-modal learning.

        \textbf{Asset Perception Datasets.} Datasets such as TT100K \citep{zhu2016traffic}, GTSDB \citep{houben2013detection}, and ARTSv2 \citep{wilson2022object} focus on detecting specific facilities but are ill-suited for systematic inventory. They are predominantly limited to single-modality imagery, lacking the 3D depth information needed to map assets to real-world coordinates. Crucially, they typically lack \textbf{cross-frame tracking IDs}, making it impossible to perform asset counting or deduplication. Furthermore, while offering basic category labels, these datasets fail to capture \textbf{attributes and statuses} essential for maintenance---such as attributes (e.g., material, shape) and statuses including visual defects (e.g., occlusion), physical damage (e.g., breakage), and functional anomalies (e.g., malfunctioning lights).
        
        In conclusion, existing benchmarks exhibit a clear disconnect between perception capabilities and management requirements. None of them simultaneously provide multi-modal data, precise 2D-3D instance mapping, cross-frame tracking, and comprehensive attribute and status descriptions. To address these limitations, we propose \textbf{WHU-Infra3D}, a unified platform with full-stack annotations designed to bridge the gap between computer vision and intelligent infrastructure management.
    
    \subsection{Methods for Infrastructure Inventory}
        Automated infrastructure inventory is a systematic process that transforms raw sensor data into a structured asset database \citep{liu2025training}. This pipeline typically involves a series of interconnected tasks: identifying targets from sensory inputs, associating observations across time and space, determining their precise 3D locations, delineating their geometric boundaries, and finally, diagnosing their attributes and statuses. Accordingly, we review the related methodologies in five key aspects: (1) 2D Detection, which serves as the foundation for target discovery; (2) 2D Cross-View Matching, which ensures consistent identity across multiple views; (3) 3D Geo-Identification, which recovers the precise geospatial coordinates of assets; (4) 3D Point Cloud Segmentation, which extracts detailed 3D geometry; and (5) Multi-Modal Attribute Recognition, which provides the cognitive understanding essential for maintenance decision-making. It is worth noting that 3D Geo-Identification and 3D Point Cloud Segmentation are often parallel or alternative steps, depending on the specific application requirements (e.g., coordinate mapping vs. 3D reconstruction) and the availability of data modalities (e.g., image-only vs. LiDAR-integrated).

        \textbf{2D Detection.} Accurate 2D perception from street-level imagery serves as the foundation for downstream inventory tasks. While classical closed-set detectors like YOLO \citep{redmon2016you} and SERNet \citep{erisen2024sernet} have achieved robust performance \citep{liu2016ssd,jain2023oneformer}, they are constrained by fixed category definitions, necessitating costly re-annotation for emerging infrastructure types \citep{liu2025training}. To overcome this rigidity, Open-Vocabulary Detection (OVD) has emerged as a flexible alternative. Models such as GLIP \citep{li2022grounded}, Grounding DINO \citep{liu2024grounding,ren2024grounding}, and DINO-X \citep{ren2024dino} leverage pre-trained vision-language alignment to detect arbitrary categories via text prompts or visual prompts \citep{jiang2024t}, while foundation segmentation models like SAM \citep{kirillov2023segment,ravi2024sam,carion2025sam} further enable high-quality instance segmentation from simple box prompts. However, since these foundation models are primarily trained on general-purpose data, their zero-shot performance often degrades on domain-specific infrastructure (e.g., surveillance cameras) due to significant domain gaps. This uncertainty underscores the necessity of a dedicated benchmark to systematically evaluate their effectiveness in real-world inventory scenarios.

        \textbf{2D Cross-View Matching.} Associating infrastructure observations across sparse views is critical for asset counting and deduplication. Unlike video-based Multi-Object Tracking (MOT), street-view imagery typically features extremely wide baselines, causing drastic appearance changes that render traditional smoothness-based trackers (e.g., DeepSORT \citep{wojke2017simple}, TrackFormer \citep{meinhardt2022trackformer}) ineffective. Consequently, existing approaches treat this as a cross-view matching problem, categorized into three streams \citep{liu2025training}. \textbf{Geometry-based methods} \citep{krylov2018automatic, nassar2019simultaneous} rely on ray intersection constraints but often fail in dense urban corridors where object sparsity assumptions are violated. \textbf{Graph-based methods} \citep{nassar2020geograph} model detections as nodes and consistency as edges, yet their reasoning is brittle to spurious connections. \textbf{Similarity learning methods} \citep{chaabane2021end, wilson2022object} directly predict pairwise affinity by fusing visual and geometric embeddings. However, they are fundamentally limited to local frame-to-frame interactions, where sequential chaining leads to error accumulation. To address these limitations, recent works \citep{liu2026SSVI-3D} have introduced spatial-attention mechanisms to model global geometric relationships among multi-view observations, enabling robust long-range association without relying on unreliable appearance cues.

        \textbf{3D Geo-Identification.} Anchoring observations to real-world coordinates is typically achieved via two paradigms. \textbf{Single-view estimation methods} \citep{campbell2019detecting, wilson2022object} rely on depth prediction or regression from individual images, subsequently fusing results via heuristics like averaging. While flexible, they often underutilize multi-view constraints, limiting precision. In contrast, \textbf{Multi-view geometric methods} \citep{krylov2018automatic, nassar2019simultaneous} leverage triangulation across calibrated frames to recover 3D positions consistent with physical constraints. Although offering higher theoretical accuracy, standard triangulation requires precise camera calibration and is sensitive to feature matching errors across sparse views.

        \textbf{3D Point Cloud Segmentation.} Extracting detailed 3D geometry from laser scanning data is essential for asset modeling. Early methods relied on hand-crafted geometric features (e.g., supervoxels, region growing) \citep{wang2021pole, li2022supervoxel} or classical machine learning classifiers like SVMs \citep{li2019semantic, truong2022efficient} to extract specific targets such as poles and traffic signs. However, these approaches struggle to generalize to diverse infrastructure categories due to limited feature descriptiveness. With the advent of deep learning, point-based networks \citep{qi2017pointnet++, thomas2019kpconv} and voxel-based architectures \citep{choy20194d} have advanced semantic and instance segmentation \citep{wu2024point}. Nevertheless, pure point cloud methods are inherently limited by sparse resolution and lack of texture, making it difficult to distinguish objects with similar geometry but different semantics.
        To address this, multi-modal methods \citep{qi2018frustum, zhou2022street, gong2020frustum} leverage 2D detectors to guide 3D segmentation, lifting image proposals into 3D frustums for precise boundary delineation. While effective, these approaches typically operate under closed-set assumptions and rely on single-frame frustums, often underutilizing multi-view observations. Recent training-free frameworks \citep{liu2025training} attempt to overcome these limitations by projecting open-set 2D masks from multiple views onto point clouds, offering a more flexible solution for diverse infrastructure inventory.
    
        \textbf{Multi-Modal Attribute Recognition.} Beyond localization, parsing attributes and statuses is pivotal for automated maintenance \citep{ma2022computer}. Traditional approaches rely on rigid, supervised pipelines for isolated targets \citep{behrendt2017deep, aygun2024building}, which suffer from limited scalability and fail to address long-tailed statuses \citep{tabernik2019deep}. The emergence of Vision-Language Models (VLMs) offers a flexible alternative. Early domain applications utilized contrastive models like CLIP to achieve open-set attribute matching without training \citep{liu2025training, radford2021learning}. More recently, generative LVLMs (e.g., Qwen-VL \citep{bai2023qwen}) have been leveraged as cognitive backbones in training-free agent frameworks to infer complex statuses through direct multi-modal prompting \citep{liu2026SSVI-3D}. However, to mitigate hallucinations and ensure professional compliance, the latest frameworks further integrate domain-specific fine-tuning with Retrieval-Augmented Generation (RAG), transforming generic models into schema-conformant experts for engineering workflows \citep{Fu2026}.
    

    \begin{figure}[htbp]
        \centering{
        \includegraphics[width=0.50\textwidth]{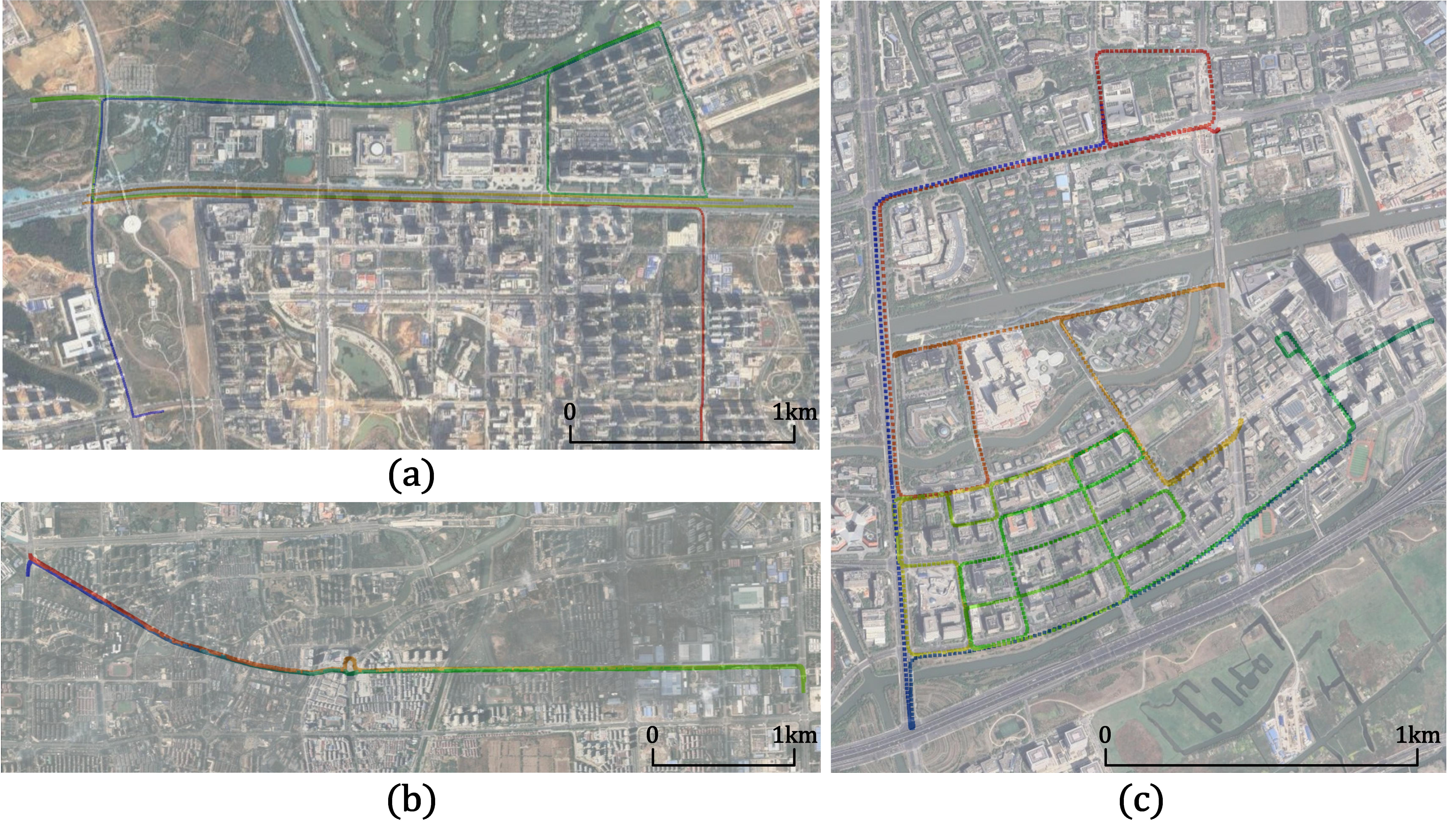}}%
        \caption{Overview of the data collection trajectories across three cities. The panels illustrate the mobile mapping routes in (a) Wuhan, (b) Nanjing, and (c) Shanghai.}
        \label{fig:trajectory}
    \end{figure}

    \begin{figure*}[t]
        \centering
        \includegraphics[width=\textwidth]{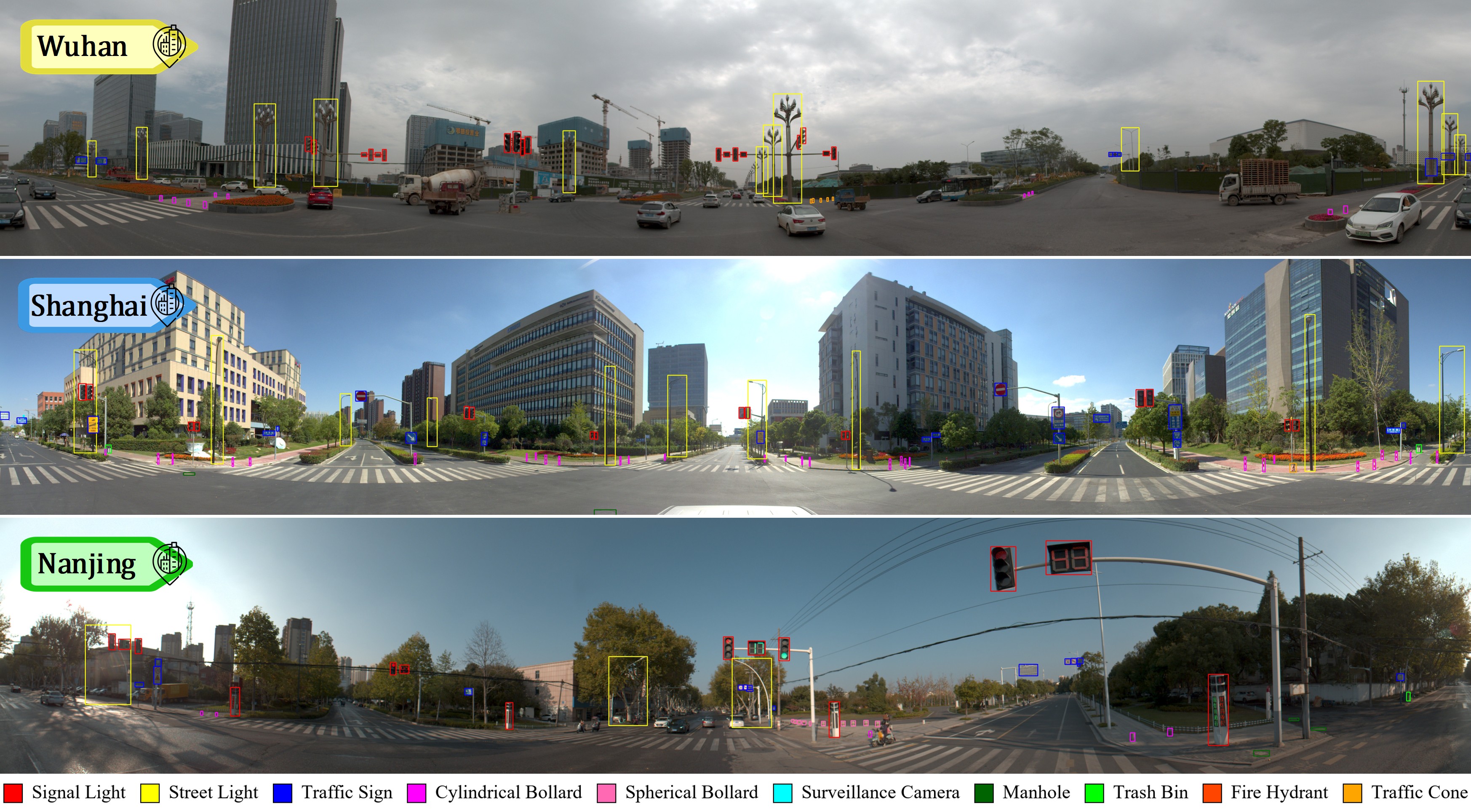}
        \caption{Representative 2D bounding box annotations on panoramic images from Wuhan, Shanghai, and Nanjing. The diverse urban layouts, lighting conditions, and infrastructure appearances across these three cities highlight the dataset's scale and complexity. Note that the images are vertically cropped to focus on the road scenes for better visualization.}
        \label{fig:2d_annotations}
    \end{figure*}

    \section{The WHU-Infra3D Dataset}
    \label{sec:dataset}

    This section presents WHU-Infra3D from three complementary perspectives: data acquisition, full-stack annotation design, and statistical characteristics. We first describe the sensing platforms and collection protocol, then detail the multi-dimensional annotation schema, and finally summarize the dataset scale and distribution properties that motivate the downstream benchmarks.
    
    \subsection{Data Acquisition}
    \label{subsec:acquisition}
    The WHU-Infra3D dataset is collected by Mobile Mapping Systems (MMS) navigating across three major Chinese cities: Shanghai, Wuhan, and Nanjing. To capture comprehensive environmental context, these platforms are equipped with high-precision GNSS/IMU modules for robust localization, high-frequency LiDAR scanners for dense 3D point cloud generation, and high-resolution panoramic cameras for visual texture acquisition.
    
    To encompass diverse sensing conditions and temporal variations, the dataset spans multiple years and employs distinct MMS platforms: AS-900HL for Shanghai (2018), HiScan-Z for Wuhan (2021), and KZH1750 for Nanjing (2024). The detailed parameters of these laser scanning systems are listed in Table~\ref{tab:sensor_params}. The data collection trajectories for each city are visualized in Fig. \ref{fig:trajectory}.

    \begin{table}[htbp]
        \centering
        \caption{Parameters of the laser scanning systems.}
        \label{tab:sensor_params}
        \resizebox{\linewidth}{!}{
        \begin{tabular}{lccc}
        \toprule
        Parameter & AS-900HL & HiScan-Z & KZH1750 \\ \midrule
        Maximum scanning distance & 920 m & 119 m & 1500 m \\
        Laser emission frequency & 550 k pts/s & 1010 k pts/s & 550 k pts/s \\
        Scanning frequency & 200 Hz & 200 Hz & 200 Hz \\
        Panoramic camera resolution & 8192 $\times$ 4096 & 8192 $\times$ 4096 & 8192 $\times$ 4096 \\ \bottomrule
        \end{tabular}
        }
    \end{table}

    \subsection{Data Annotation}
    \label{subsec:data_annotation}
        
        To facilitate a comprehensive spectrum of tasks ranging from foundational object detection to advanced urban asset management, we constitute a highly structured, multi-dimensional annotation framework for WHU-Infra3D. The dataset systematically focuses on 10 critical categories of roadside infrastructure: Traffic Sign (Tra. Sign), Street Light (Str. Light), Signal Light (Sig. Light), Surveillance Camera (Surv. Cam.), Cylindrical Bollard (Cyl. Bol.), Fire Hydrant (Hydrant), Trash Bin (Trash Bin), Manhole (Manhole), Traffic Cone (Tra. Cone), and Spherical Bollard (Sph. Bol.). This selection covers a wide range of structural types and functions, including vertical pole-like objects (e.g., street lights), ground-level facilities (e.g., manholes), fixed assets (e.g., trash bins), and temporary objects (e.g., traffic cones). Consequently, the hierarchical annotation encompasses four key dimensions:

        \textbf{2D Detection Annotations.} Spanning the entire 53.8 km trajectory, we establish exhaustive 2D bounding box annotations on the visual data. The native panoramic images were systematically unrolled into rectilinear sub-views to conform to standard object detection paradigms. Initial proposals were generated via a semi-automated pipeline, followed by rigorous, multi-stage manual refinement to guarantee high localization quality. As illustrated in Figure \ref{fig:2d_annotations}, our annotations cover diverse urban scenarios across three major cities—Wuhan, Shanghai, and Nanjing. The data naturally captures infrastructure under varying lighting conditions, different object densities, and frequent occlusions. This cross-city coverage introduces realistic domain shifts in infrastructure appearance, providing a challenging benchmark for evaluating the generalization capability of contemporary perception models.

    \begin{figure}[!t]
        \centering
        \includegraphics[width=\linewidth]{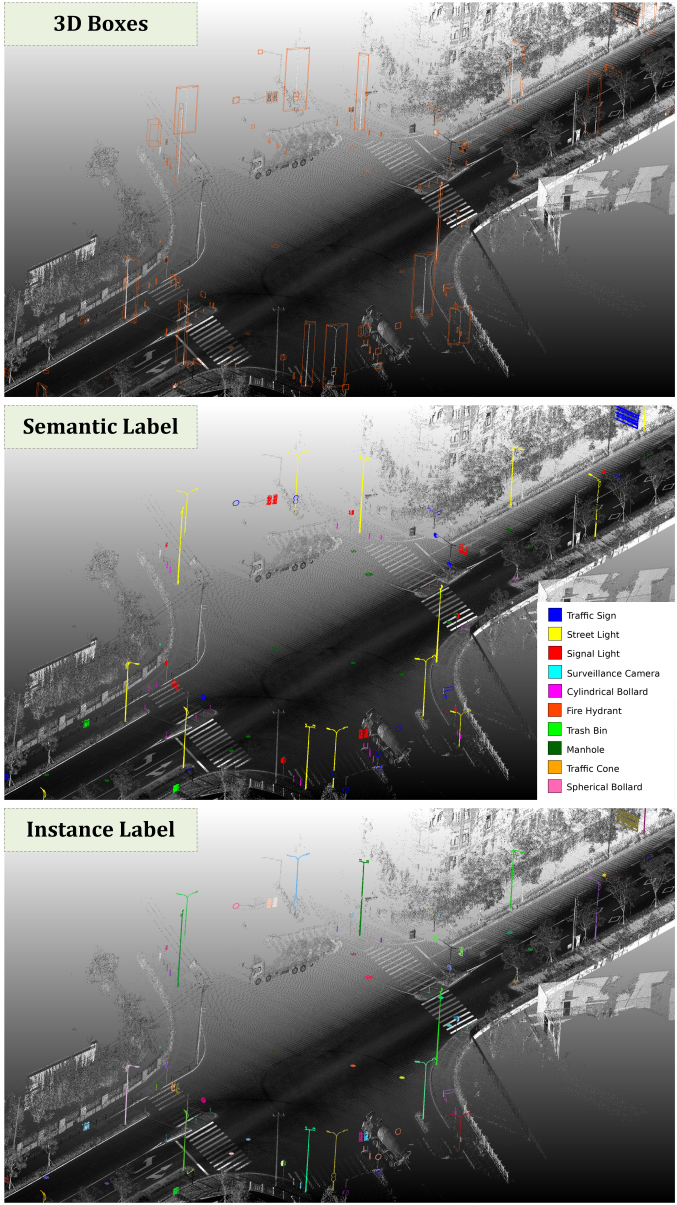}
        \caption{Representative visualization of 3D annotations in WHU-Infra3D. From top to bottom: 3D oriented bounding boxes, semantic labels with class-color legend, and instance labels.}
        \label{fig:3d_annotation_overview}
    \end{figure}

        \begin{figure}[!t]
            \centering
            \includegraphics[width=\linewidth]{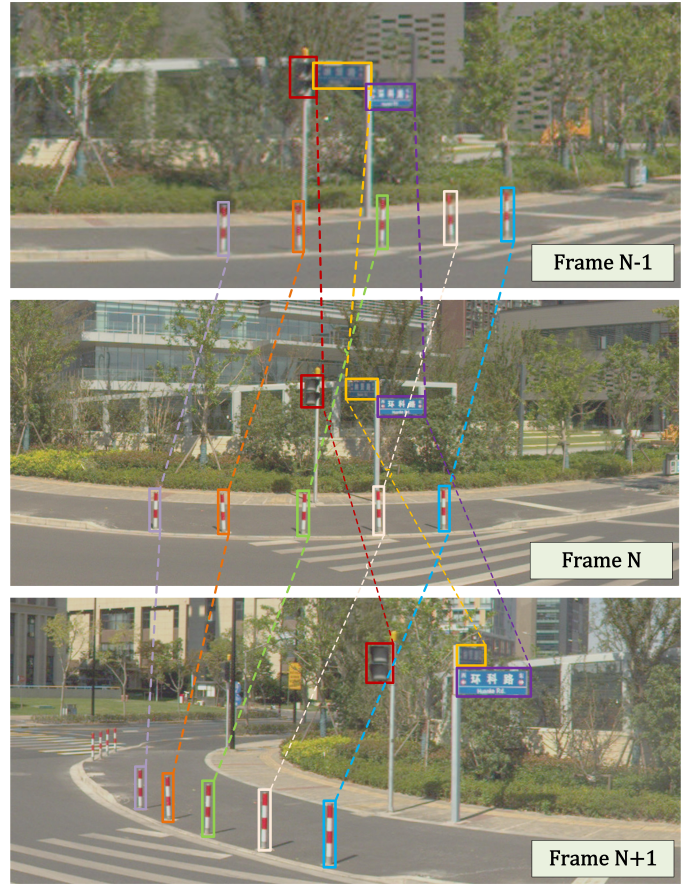}
            \caption{Illustration of cross-frame instance association in WHU-Infra3D. The colored dashed lines connect the same physical objects (e.g., traffic signs, signal lights, bollards) across sequential panoramic frames (from Frame N-1 to Frame N+1). Note that the images are cropped for better visualization.}
            \label{fig:cross_frame}
        \end{figure}

        \textbf{3D Perception Annotations.} We provide dense 3D annotations within the LiDAR point clouds for a core subset covering approximately 10 km. These annotations encompass point-wise semantic labels, instance masks, and 3D center points for each infrastructure asset. Additionally, we provide 3D oriented bounding boxes (OBB) that precisely delineate their spatial extent and orientation. Figure \ref{fig:3d_annotation_overview} illustrates a representative example featuring 3D bounding boxes alongside semantic and instance labels.

        \textbf{Instance Association Annotations.} We assign a \textbf{globally unique instance ID} to each physical infrastructure object. This ID system is crucial not only for accurate deduplication to count the exact number of unique instances, but also for aggregating multi-view observations to obtain a more comprehensive representation of an object's overall attributes and statuses. This association is established in two aspects: (1) \textbf{Cross-frame Association}, as visualized in Figure \ref{fig:cross_frame}, where the same object retains its unique ID across sequential image frames, enabling robust tracking; and (2) \textbf{Cross-modal Association}, where the 2D image bounding box and the corresponding 3D point cloud instance share an identical ID, establishing a precise one-to-one mapping between the visual and geometric modalities.

\begin{figure}[!t]
            \centering{
            \includegraphics[width=\linewidth]{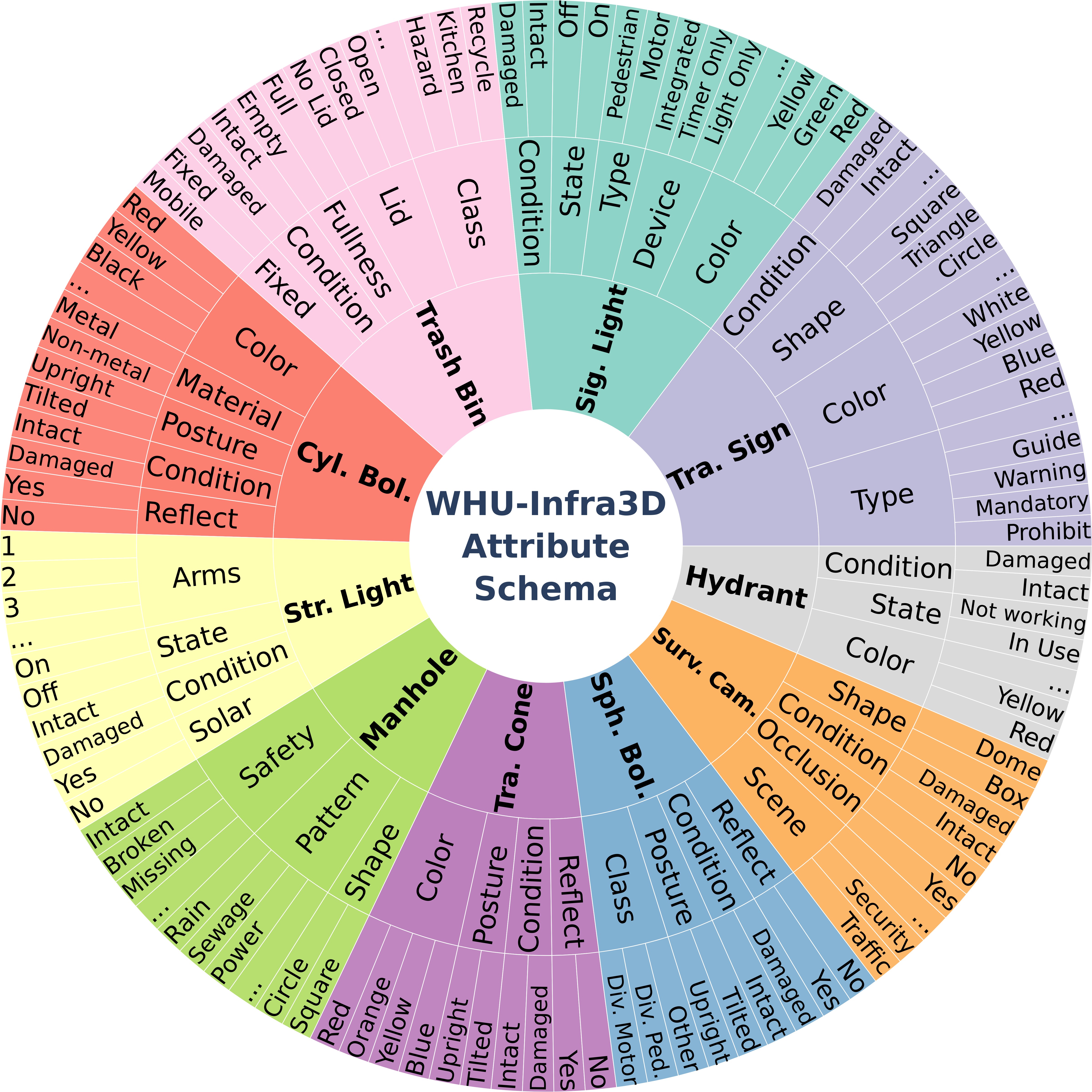}}%
            \caption{Visualization of the hierarchical attribute schema in WHU-Infra3D. The concentric rings represent object categories, attributes, and specific values from the center outwards.}
            \label{fig:attr_schema}
        \end{figure}

        \textbf{Attribute and Status Annotations.} To bridge the gap between perceptual data and physical infrastructure management, we developed a comprehensive attribute schema designed for fine-grained asset inventory. As detailed in Figure \ref{fig:attr_schema}, this schema extends beyond basic visual attributes to encapsulate critical \textbf{functional and safety statuses}. For example, in addition to attributes like \textit{Shape} and \textit{Material}, we annotated statuses. These include the \textit{Working Status} of traffic lights (e.g., operational vs. malfunctioning), the \textit{Fullness} of trash receptacles, and the \textit{Posture} of bollards (e.g., upright vs. tilted). These rich semantic annotations translate raw sensory data into actionable insights for infrastructure maintenance.

        To address the varying annotation costs, we structured the dataset into two mutually exclusive tiers: the \textbf{Base Set} and the \textbf{Core Set}. The \textbf{Base Set} encompasses 44.7 km of the trajectory and provides exclusively 2D detection annotations, serving as a large-scale foundation for visual detection training. The \textbf{Core Set}, a high-quality subset spanning the remaining 9.1 km curated from diverse environmental scenarios, is enriched with all four dimensions of annotations. This full-stack subset serves as a paramount benchmark for fine-grained asset inventory and multi-modal learning.

    \subsection{Dataset Statistics}
    \label{subsec:statistics}

    Table~\ref{tab:dataset_stats} summarizes the scale of WHU-Infra3D across the three surveyed cities under the two-tier annotation protocol described in Section~\ref{subsec:data_annotation}. The \textbf{Total} columns aggregate the entire 53.8 km trajectory (including both Base Set and Core Set), providing complete 2D box annotations. The \textbf{Core Set} columns detail the high-quality subset with full-stack annotations, including 2D boxes, 3D instance annotations, and attribute and status labels.

    \begin{table}[!t]
        \centering
        \caption{Dataset statistics of WHU-Infra3D.}
        \label{tab:dataset_stats}
        \resizebox{0.50\textwidth}{!}{%
        \begin{tabular}{lccc|cccc}
        \toprule
        \multirow{2}{*}{\textbf{City}} & \multicolumn{3}{c|}{\textbf{Total}} & \multicolumn{4}{c}{\textbf{Core Set}} \\
        \cmidrule(lr){2-4} \cmidrule(lr){5-8}
         & \textbf{Traj. (km)} & \textbf{Images} & \textbf{2D Box} & \textbf{Images} & \textbf{2D Box} & \textbf{3D Inst.} & \textbf{Attr./Status} \\
        \midrule
        Wuhan & 20.15 & 1,975 & 59,069 & 229 & 7,311 & 973 & 31,068 \\
        Shanghai & 21.81 & 1,553 & 64,638 & 260 & 11,785 & 993 & 50,797 \\
        Nanjing & 11.84 & 1,921 & 51,314 & 615 & 23,664 & 1,233 & 99,486 \\
        \midrule
        Total & 53.80 & 5,449 & 175,021 & 1,104 & 42,760 & 3,199 & 181,351 \\
        \bottomrule
        \end{tabular}%
        }
    \end{table}

\begin{figure}[!t]
        \centering
        \includegraphics[width=\columnwidth]{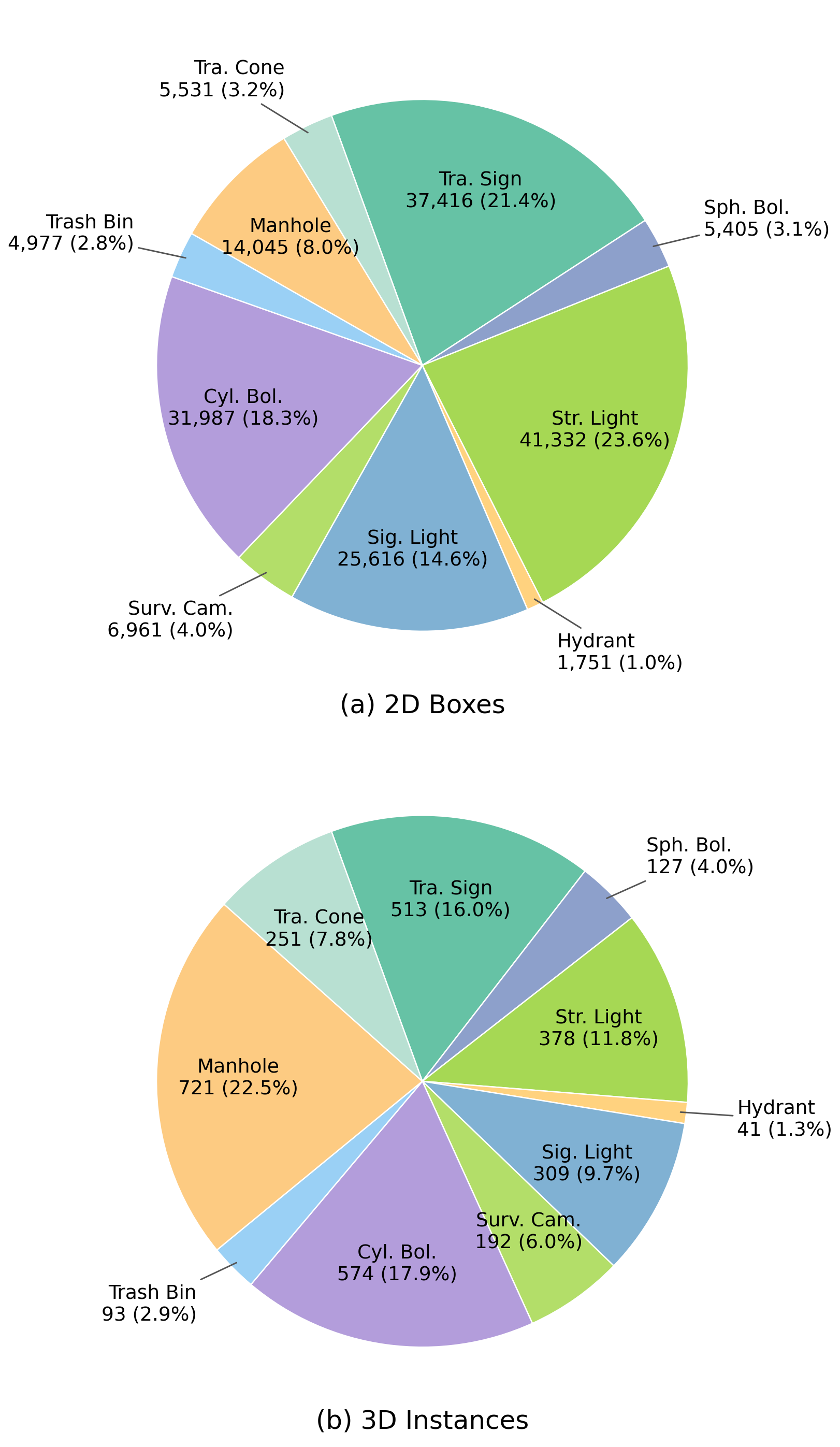}
        \caption{Category distribution of WHU-Infra3D in 2D box annotations (a) and 3D instance annotations (b). Both modalities exhibit a pronounced long-tail imbalance.}
        \label{fig:category_pies}
    \end{figure}

    Figure~\ref{fig:category_pies} illustrates the pronounced long-tail category distributions across both 2D and 3D instance annotations. In both modalities, Street Light, Traffic Sign, and Cylindrical Bollard constitute the majority of instances, whereas Fire Hydrant, Trash Bin, and Spherical Bollard appear comparatively rarely. Beyond the long-tail pattern, a clear cross-modal discrepancy can be observed: Manhole accounts for a much larger share in 3D instances but a smaller share in 2D boxes, mainly because manholes are ground-level and geometrically compact, and their 2D appearance is often weakened by perspective foreshortening and frequent occlusion from passing vehicles. Street Light shows the opposite trend in 2D observations, as its tall vertical structure is highly salient and remains visible over long image ranges. Consequently, the average number of 2D observations per 3D instance is highly imbalanced across categories, which poses additional challenges for downstream cross-view matching and 3D localization, where robust performance is required under highly varying observation availability.

    \subsection{Dataset Characteristics and Advantages}
    The WHU-Infra3D dataset presents several distinct characteristics and advantages that address critical gaps in current infrastructure inventory research:
    
    \begin{itemize}
        \item \textbf{Full-stack and Multi-modal Synergy:} Unlike previous datasets that isolate specific perception tasks, WHU-Infra3D distinguishes itself by offering a full-stack benchmark that covers the entire automated infrastructure inventory workflow. Empowered by the deep integration of high-resolution panoramic images and dense 3D point clouds, it seamlessly supports a complete continuous pipeline—from 2D instance detection and cross-view matching to 3D spatial geo-identification, fine-grained point cloud segmentation, and attribute recognition. This full-stack paradigm ensures that algorithms can be comprehensively evaluated from raw multi-modal inputs to end-to-end cognitive outputs within a single unified platform.
        
        \item \textbf{High Diversity and Real-world Complexity:} Ranging across three major Chinese metropolises (Wuhan, Nanjing, and Shanghai), the dataset captures a vast array of typical urban scenarios. These explicitly include urban arterial roads, tree-lined paths, complex intersections, and multi-level overpasses. Such profound geographical and structural diversity—compounded by illumination changes and real-world occlusions—serves as a robust testbed for evaluating the open-world generalization capabilities of perception models.
        
        \item \textbf{Fine-grained Attributes and Statuses for Asset Management:} Moving beyond the conventional provision of basic bounding boxes or categories, WHU-Infra3D delivers hierarchical annotations of fine-grained attributes and statuses (e.g., detailed forms of traffic lights, sign semantics, and asset integrity/damage). These granular details intimately align with the demanding practical requirements of real-world road asset maintenance and smart city systems, shifting the research frontier from rudimentary object localization to deep situational understanding.
    \end{itemize}

    \begin{table*}[t]
        \centering
        \caption{Summary of benchmark tasks defined in WHU-Infra3D.}
        \label{tab:tasks}
        \resizebox{\textwidth}{!}{%
        \begin{tabular}{l:l:l:l}
        \toprule
        \textbf{Task} & \textbf{Input} & \textbf{Output} & \textbf{Metric} \\ \midrule
        2D Infrastructure Detection & Panoramic Images & 2D Boxes + Category & mAP, mAP@50 \\
        2D Cross-view Matching & Image Sequence + Poses & Unique IDs (Cross-view Associations) & Homogeneity, Completeness \\
        3D Geo-identification & Matched Observations + Poses & 3D Coordinates + Geo-verified IDs & Identification Precision@1m, Identification Recall@1m, Localization Error \\
        Point Cloud Segmentation & Point Cloud + Images & Point-wise Instance Masks & Segmentation Precision@IoU0.5, Segmentation Recall@IoU0.5, mIoU \\
        Attribute Recognition & Object ROI (2D/3D) & Structured Attributes & Accuracy \\ \bottomrule
        \end{tabular}%
        }
    \end{table*}
    
    \section{Benchmark Tasks and Baselines}

    This section establishes the benchmark protocols and reports baseline results on five core tasks for infrastructure inventory. We first define each task and its evaluation metrics, and then provide quantitative analyses covering in-domain performance, cross-city generalization, geometric localization, 3D segmentation, and attribute recognition.

    \subsection{Task Definitions}
        To systematically evaluate the capabilities of existing algorithms in infrastructure inventory, we define five core benchmark tasks, as summarized in Table~\ref{tab:tasks}.
        
        \textbf{Task 1: 2D Infrastructure Detection.} Serves as the foundation, requiring models to detect and classify objects in panoramic images, evaluated by \textit{mAP} and \textit{mAP@50}.
        
        \textbf{Task 2: 2D Cross-view Matching.} Focuses on cross-frame association in image space. Models must link observations of the same physical asset across sparse views. We evaluate this task using \textit{Homogeneity} and \textit{Completeness}: Homogeneity measures cluster purity (whether each predicted cluster contains observations from only one asset), while Completeness measures cluster integrity (whether observations of one asset are grouped into the same cluster).
        
        \textbf{Task 3: 3D Geo-identification.} Focuses on location-aware identity confirmation. Given cross-view correspondences and calibrated poses, models estimate asset-level 3D coordinates. Evaluated using \textit{Identification Precision@1m}, \textit{Identification Recall@1m}, and \textit{Localization Error}. Predictions are matched with ground truth objects based on correct identity association and Euclidean localization error $\leq 1$m, from which Precision, Recall, and the mean Localization Error of true positives are calculated.
        
        \textbf{Task 4: Point Cloud Segmentation.} Focuses on geometric delineation, requiring the segmentation of 3D instance masks from LiDAR point clouds, evaluated by \textit{Segmentation Precision@IoU0.5}, \textit{Segmentation Recall@IoU0.5}, and \textit{mIoU}. Predictions are assigned to ground-truth instances of the same category using a one-to-one mapping with an IoU threshold of $0.5$, from which Precision and Recall are computed. mIoU is reported as the category-averaged IoU over matched pairs.
        
        \textbf{Task 5: Attribute Recognition.} Targets cognitive understanding, where models must diagnose attributes and statuses from object ROIs, evaluated by the mean classification accuracy across all attribute categories.

    \begin{table*}[t]
        \centering
        \caption{In-domain evaluation for 2D Infrastructure Detection. Models are trained on the Base Set and evaluated on the Core Set across all three cities combined. Category abbreviations are defined in Section~\ref{subsec:data_annotation}.}
        \label{tab:det_indomain}
        \resizebox{\textwidth}{!}{%
        \begin{tabular}{l:cc:cccccccccc}
        \toprule
        \multirow{2}{*}{\textbf{Method}} & \multicolumn{2}{c:}{\textbf{Mean}} & \multicolumn{10}{c}{\textbf{Per-class AP@50}} \\ \cmidrule(lr){2-3} \cmidrule(lr){4-13}
         & \textbf{mAP} & \textbf{mAP@50} & \textbf{Tra. Sign} & \textbf{Str. Light} & \textbf{Sig. Light} & \textbf{Surv. Cam.} & \textbf{Cyl. Bol.} & \textbf{Hydrant} & \textbf{Trash Bin} & \textbf{Manhole} & \textbf{Tra. Cone} & \textbf{Sph. Bol.} \\ \midrule
        Faster R-CNN$^\dagger$ & 0.335 & 0.566 & 0.607 & 0.529 & 0.659 & 0.353 & 0.620 & 0.555 & 0.675 & 0.427 & 0.680 & 0.554 \\
        DINO$^\dagger$ & 0.471 & 0.720 & 0.719 & 0.709 & 0.791 & 0.602 & 0.778 & 0.726 & 0.774 & 0.619 & 0.755 & 0.724 \\
        YOLOv26-M$^\dagger$ & 0.537 & 0.753 & 0.773 & 0.717 & 0.834 & 0.635 & 0.794 & 0.775 & 0.823 & 0.629 & 0.815 & 0.739 \\ \hdashline
        GLIP$^\ddagger$ & 0.175 & 0.289 & 0.382 & 0.276 & 0.451 & 0.118 & 0.143 & 0.423 & 0.341 & 0.172 & 0.378 & 0.088 \\
        YOLO-World$^\ddagger$ & 0.223 & 0.325 & 0.475 & 0.154 & 0.534 & 0.225 & 0.080 & 0.578 & 0.424 & 0.301 & 0.480 & 0.003 \\
        Grounding DINO$^\ddagger$ & 0.228 & 0.361 & 0.496 & 0.417 & 0.658 & 0.165 & 0.16 & 0.533 & 0.443 & 0.210 & 0.488 & 0.041 \\
        \textbf{FT-G-DINO}$^\ddagger$ & \textbf{0.545} & \textbf{0.865} & \textbf{0.890} & \textbf{0.869} & \textbf{0.905} & \textbf{0.747} & \textbf{0.880} & \textbf{0.894} & \textbf{0.881} & \textbf{0.819} & \textbf{0.853} & \textbf{0.908} \\ \bottomrule
        \end{tabular}%
        }
    \end{table*}

    \begin{table*}[t]
        \centering
        \caption{Cross-city generalization analysis for 2D Infrastructure Detection (Leave-One-City-Out). Models are trained on the Base Set from two source cities and tested on the Core Set of the unseen third city. We compare the representative closed-set detector (Closed) against our open-set baseline (Open). Category abbreviations are defined in Section~\ref{subsec:data_annotation}.}
        \label{tab:det_crosscity}
        \resizebox{\textwidth}{!}{%
        \begin{tabular}{l:cc:cccccccccc}
        \toprule
        \multirow{2}{*}{\textbf{Method (Test City)}} & \multicolumn{2}{c:}{\textbf{Mean}} & \multicolumn{10}{c}{\textbf{Per-class AP@50}} \\ \cmidrule(lr){2-3} \cmidrule(lr){4-13}
         & \textbf{mAP} & \textbf{mAP@50} & \textbf{Tra. Sign} & \textbf{Str. Light} & \textbf{Sig. Light} & \textbf{Surv. Cam.} & \textbf{Cyl. Bol.} & \textbf{Hydrant} & \textbf{Trash Bin} & \textbf{Manhole} & \textbf{Tra. Cone} & \textbf{Sph. Bol.} \\ \midrule
        YOLOv26-M (Wuhan)       & 0.503 & 0.696 & 0.769 & 0.608 & 0.806 & 0.548 & 0.678 & 0.825 & 0.754 & 0.356 & 0.841 & 0.776 \\
        FT-G-DINO (Wuhan)    & 0.516 & 0.742 & 0.760 & 0.664 & 0.778 & 0.618 & 0.686 & 0.815 & 0.830 & 0.626 & 0.845 & 0.799 \\ \hdashline
        YOLOv26-M (Shanghai)    & 0.354 & 0.546 & 0.626 & 0.628 & 0.725 & 0.204 & 0.500 & 0.628 & 0.599 & 0.281 & 0.652 & 0.705 \\
        FT-G-DINO (Shanghai) & 0.360 & 0.585 & 0.644 & 0.709 & 0.727 & 0.182 & 0.519 & 0.717 & 0.577 & 0.385 & 0.633 & 0.760 \\ \hdashline
        YOLOv26-M (Nanjing)     & 0.293 & 0.444 & 0.445 & 0.376 & 0.365 & 0.413 & 0.455 & 0.382 & 0.558 & 0.367 & 0.606 & 0.477 \\
        FT-G-DINO (Nanjing)  & 0.309 & 0.538 & 0.530 & 0.444 & 0.505 & 0.440 & 0.487 & 0.658 & 0.564 & 0.495 & 0.621 & 0.634 \\ 
        \bottomrule
        \end{tabular}%
        }
    \end{table*}

    \begin{table*}[t]
        \centering
        \caption{Quantitative benchmark results under the Cross-City Transfer setting for \textbf{Task 2} (2D Cross-view Matching) and \textbf{Task 3} (3D Geo-identification). The clustering metrics (Homogeneity and Completeness) are reported from the GT-based association benchmark. Identification precision/recall are reported as Precision@1m and Recall@1m, where TP requires both correct identity and localization error $<1$ m; localization error is the mean Euclidean error over TP pairs.}
        \label{tab:loc_results_final}
        
        \resizebox{\textwidth}{!}{%
        \begin{tabular}{l : *{5}{w{c}{1.1cm}} : *{5}{w{c}{1.1cm}} : *{5}{w{c}{1.1cm}}}
        \hline
        
        \multicolumn{1}{l}{\multirow{3}{*}{\textbf{Method / Category}}} & 
        \multicolumn{5}{c}{\textbf{Wuhan}} & 
        \multicolumn{5}{c}{\textbf{Shanghai}} & 
        \multicolumn{5}{c}{\textbf{Nanjing}} \\
        \cmidrule(lr){2-6} \cmidrule(lr){7-11} \cmidrule(lr){12-16}
        
        \multicolumn{1}{l}{} & 
        \multicolumn{2}{c}{\textit{Clus.}} & \multicolumn{2}{c}{\textit{Ident.}} & \multicolumn{1}{c}{\textit{Loc.}} & 
        \multicolumn{2}{c}{\textit{Clus.}} & \multicolumn{2}{c}{\textit{Ident.}} & \multicolumn{1}{c}{\textit{Loc.}} & 
        \multicolumn{2}{c}{\textit{Clus.}} & \multicolumn{2}{c}{\textit{Ident.}} & \textit{Loc.} \\
        \cmidrule(lr){2-3} \cmidrule(lr){4-5} \cmidrule(lr){6-6} 
        \cmidrule(lr){7-8} \cmidrule(lr){9-10} \cmidrule(lr){11-11} 
        \cmidrule(lr){12-13} \cmidrule(lr){14-15} \cmidrule(lr){16-16}
        
        \multicolumn{1}{l}{} & 
        \textbf{Hom.} & \textbf{Cmp.} & \textbf{Pre.} & \textbf{Rec.} & \multicolumn{1}{c}{\textbf{Err.}} & 
        \textbf{Hom.} & \textbf{Cmp.} & \textbf{Pre.} & \textbf{Rec.} & \multicolumn{1}{c}{\textbf{Err.}} & 
        \textbf{Hom.} & \textbf{Cmp.} & \textbf{Pre.} & \textbf{Rec.} & \textbf{Err.} \\
        \hline
        
        \multicolumn{16}{c}{\textbf{\textit{Method Comparison (Average over all classes)}}} \\
        \hline

        Visual-cue & 43.0 & 73.9 & 35.2 & 21.7 & 0.20 & 36.0 & 65.9 & 23.9 & 12.6 & 0.28 & 41.3 & 65.1 & 24.2 & 15.3 & 0.28 \\
        Visual-cue\textsuperscript{R} & 98.3 & 94.4 & 85.0 & 75.9 & 0.11 & 97.6 & 93.6 & 77.0 & 79.2 & 0.15 & 98.5 & 95.5 & 67.5 & 64.1 & 0.17 \\
        \hdashline
        Geometry-cue & 90.8 & 86.6 & 57.5 & 68.4 & 0.13 & 88.8 & 81.3 & 39.9 & 59.0 & 0.20 & 88.1 & 80.9 & 41.3 & 58.6 & 0.20 \\
        Geometry-cue\textsuperscript{R} & 98.5 & 95.8 & 85.2 & 75.9 & 0.11 & 98.1 & 95.1 & 77.3 & 78.5 & 0.15 & 98.5 & 96.7 & 69.7 & 63.4 & 0.17 \\
        \hdashline
        SVII-3D & 96.5 & 88.6 & 62.5 & 76.9 & 0.14 & 94.6 & 83.7 & 43.9 & 72.9 & 0.16 & 87.6 & 83.0 & 40.1 & 56.1 & 0.18 \\
        SVII-3D\textsuperscript{R} & \textbf{99.2} & \textbf{97.9} & \textbf{90.3} & \textbf{80.9} & \textbf{0.11} & \textbf{98.8} & \textbf{97.5} & \textbf{81.2} & \textbf{79.9} & \textbf{0.14} & \textbf{98.8} & \textbf{98.0} & \textbf{70.1} & \textbf{64.0} & \textbf{0.17} \\

        \hline
        \multicolumn{16}{c}{\textbf{\textit{Per-class Performance (SVII-3D\textsuperscript{R})}}} \\
        \hline
        
        Traffic Sign & 99.6 & 97.5 & 79.4 & 87.5 & 0.10 & 99.2 & 97.7 & 85.4 & 91.6 & 0.13 & 98.9 & 97.3 & 64.3 & 78.5 & 0.12 \\
        Street Light & 99.6 & 99.4 & 96.0 & 97.9 & 0.24 & 99.0 & 97.9 & 73.5 & 93.9 & 0.27 & 98.2 & 97.3 & 80.3 & 43.8 & 0.53 \\
        Signal Light & 98.3 & 97.8 & 94.5 & 77.5 & 0.10 & 98.2 & 97.8 & 83.7 & 96.4 & 0.14 & 99.2 & 97.8 & 64.9 & 83.7 & 0.17 \\
        Surveillance Camera & 99.3 & 99.0 & 89.5 & 94.4 & 0.09 & 97.6 & 97.1 & 57.1 & 61.5 & 0.13 & 98.4 & 98.8 & 56.5 & 68.0 & 0.15 \\
        Cylindrical Bollard & 99.5 & 97.8 & 88.2 & 82.2 & 0.06 & 98.7 & 96.7 & 84.8 & 90.4 & 0.10 & 99.3 & 97.5 & 70.6 & 83.3 & 0.06 \\
        Fire Hydrant & 100.0 & 100.0 & 100.0 & 87.5 & 0.06 & 100.0 & 99.2 & 93.8 & 78.9 & 0.11 & 100.0 & 100.0 & 83.3 & 38.5 & 0.07 \\
        Trash Bin & 100.0 & 99.5 & 94.1 & 100.0 & 0.08 & 100.0 & 99.7 & 88.6 & 97.5 & 0.12 & 100.0 & 99.8 & 79.4 & 87.1 & 0.11 \\
        Manhole & 99.9 & 96.2 & 93.8 & 26.8 & 0.09 & 99.3 & 98.2 & 68.9 & 44.9 & 0.12 & 99.7 & 99.0 & 71.0 & 37.4 & 0.31 \\
        Traffic Cone & 97.2 & 94.6 & 86.9 & 78.5 & 0.15 & 99.1 & 94.2 & 85.0 & 48.6 & 0.14 & 96.3 & 98.6 & 72.2 & 50.0 & 0.06 \\
        Spherical Bollard & 98.4 & 97.2 & 80.6 & 76.3 & 0.10 & 96.6 & 96.5 & 91.3 & 95.5 & 0.14 & 97.7 & 94.3 & 58.6 & 69.5 & 0.14 \\
        \hline
        \end{tabular}%
        }
    \end{table*}
        
    \subsection{2D Infrastructure Detection}
    
        \textbf{Experimental Setup and Baselines.} Under the \textbf{In-domain} protocol, the entire Core Set is utilized as the validation set, while the \textbf{Base Set} constitutes the training set. Under the \textbf{Cross-city generalization} protocol, models are trained on the \textbf{Base Set} from two cities and tested directly on the Core Set of the unseen third city. We evaluate state-of-the-art detectors across two paradigms. Closed-set models (marked with $^\dagger$) include Faster R-CNN \citep{ren2015faster}, DINO \citep{zhang2023dino}, and YOLOv26 \citep{sapkota2026yolo26keyarchitecturalenhancements}. Open-set models (marked with $^\ddagger$) include GLIP \citep{li2022grounded}, YOLO-World \citep{cheng2024yolo}, Grounding DINO \citep{liu2024grounding}, and fine-tuned Grounding DINO (FT-G-DINO).
    
        \textbf{Results and Analysis.} 
        Table~\ref{tab:det_indomain} presents the results under the In-domain Protocol, where all three city datasets are merged for training and testing. Closed-set models like YOLOv26 typically achieve high performance because they fit the specific distribution of the combined dataset. Open-set models without fine-tuning may lag behind, but fine-tuning significantly narrows the gap. Among closed-set models, YOLOv26-M achieves the highest mAP of 0.537 and mAP@50 of 0.753, outperforming Faster R-CNN and DINO by a substantial margin. Among zero-shot open-set models, all three methods show considerably lower performance, with mAP@50 ranging from 0.289 to 0.361, reflecting the inherent difficulty of applying vision-language models to domain-specific urban infrastructure without any task-specific adaptation. Notably, FT-G-DINO achieves the best overall performance with an mAP of 0.545 and mAP@50 of 0.865, surpassing even the strongest closed-set model, which highlights the effectiveness of leveraging open-vocabulary pre-training followed by targeted fine-tuning.
        
        Table~\ref{tab:det_crosscity} details the Cross-city Protocol results using a "Leave-One-City-Out" strategy. We compare the best-performing closed-set detector against FT-G-DINO, highlighting the generalization gap. Closed-set models suffer significant performance drops on unseen cities due to domain shifts in infrastructure appearance (e.g., different traffic light designs). In contrast, FT-G-DINO maintains robust performance, demonstrating the value of open-vocabulary knowledge for inventory tasks. Across all three test cities, FT-G-DINO consistently outperforms YOLOv26-M in mAP@50, suggesting that open-vocabulary representations generalize more effectively to cities with greater appearance discrepancy from the training set. Both models experience the most severe performance degradation when tested on Nanjing, which achieves the lowest mAP@50 among all test cities, indicating that Nanjing presents the most distinct visual domain among the three cities. These results confirm that fine-tuned open-set models offer a more reliable solution for cross-city urban infrastructure detection compared to their closed-set counterparts.

        \textbf{Key Findings.} While closed-set detectors suffer severe performance drops across cities due to domain shifts, fine-tuning open-vocabulary foundation models effectively bridges this gap, establishing a robust paradigm for cross-city infrastructure detection.

    \subsection{2D Cross-view Matching and 3D Geo-identification}
        \textbf{Experimental Setup and Baselines.} The framework proposed in SVII-3D \citep{liu2026SSVI-3D} is adopted as the standard pipeline, which formulates cross-view association as a graph clustering problem to group observations into distinct instance trajectories. To strictly decouple association performance from detection errors, \textbf{Task 2} (2D Cross-view Matching) takes ground-truth 2D bounding boxes as input to test pure clustering quality. Conversely, \textbf{Task 3} (3D Geo-identification) utilizes the predicted bounding boxes from FT-G-DINO (the best model from Task 1) as input to assess the end-to-end localization performance in the wild. Since both metrics directly target the quality of the final global association, our benchmark revolves around the core affinity calculation strategies that drive the clustering. Three variants are constructed:
        (1) \textbf{Geometry-cue}: A \textbf{training-free} baseline that utilizes a heuristic to calculate affinity based on the inverse spatial distance between 3D rays.
        (2) \textbf{Visual-cue}: A \textbf{training-free} baseline utilizing a pre-trained CLIP \citep{radford2021learning} encoder, measuring the cosine similarity of visual embeddings as the affinity score.
        (3) \textbf{SVII-3D}: Utilizes a learnable spatial-attention network to capture geometric interaction patterns. This model follows the \textbf{Cross-city generalization} protocol, trained on the \textbf{Base Set} of two source cities and evaluated on the Core Set of the unseen third city.
        Finally, the impact of the \textbf{Geometry-guided Refinement (Geo. Ref)} module, which was also proposed in SVII-3D \citep{liu2026SSVI-3D}, is evaluated by applying it to all three baselines to test its universality as a plug-and-play post-processing step.
        
        \textbf{Results and Analysis.} Table~\ref{tab:loc_results_final} reports the quantitative results under the Cross-City Transfer setting. In this multi-task benchmark, we jointly evaluate the pure association capability (Homogeneity and Completeness from Task 2) and the end-to-end geo-identification capability (Identification Precision, Recall, and Localization Error from Task 3). Before geometric refinement, the purely visual association (\textbf{Visual-cue}) performs abysmally across all metrics; for instance, its Identification Precision and Recall in Wuhan are only 35.2\% and 21.7\%, indicating severe visual ambiguity across sparse panoramic views. Although \textbf{Geometry-cue} and \textbf{SVII-3D} exhibit substantially better initial clustering performance (e.g., SVII-3D achieves 96.5\% Homogeneity in Wuhan), their end-to-end identification Precision inherently suffers (62.5\%) due to the false positive detections passed from Task 1. However, once the \textbf{Geometry-guided Refinement (\textsuperscript{R})} is introduced as a post-processing step, there is a massive performance surge across all configurations. The refinement module effectively filters outlier intersections from noisy detections and merges fragmented clusters, boosting Identification Precision by roughly $+20\% {\sim} +40\%$ and Identification Recall by nearly $+10\% {\sim} +20\%$. Consequently, Localization Error significantly plummets to $0.11{\sim}0.17\text{m}$. Ultimately, the refined \textbf{SVII-3D\textsuperscript{R}} pipeline consistently achieves the best overall balance across the three cities, demonstrating the distinct advantage of combining learnable spatial reasoning with geometric refinement.
        
        \textbf{Key Findings.} Purely visual cross-view matching is inadequate for street-level images due to severe viewpoint changes. Integrating learnable spatial reasoning with strict geometric refinement is imperative for achieving robust instance tracking and sub-meter 3D localization accuracy.
    
    \begin{table*}[t]
        \centering
            \caption{Per-class 3D point cloud segmentation results of the method in \citep{liu2025training} under cross-city evaluation. For each city, we report \textbf{Segmentation Precision@IoU0.5}, \textbf{Segmentation Recall@IoU0.5}, \textbf{F1-Score}, and \textbf{IoU} (with city-level mean reported as mIoU).}
        \label{tab:pc_seg_compact}
        \resizebox{\textwidth}{!}{%
        \begin{tabular}{l cccc cccc cccc}
        \toprule
        \multirow{2}{*}{\textbf{Category}} & \multicolumn{4}{c}{\textbf{Wuhan}} & \multicolumn{4}{c}{\textbf{Shanghai}} & \multicolumn{4}{c}{\textbf{Nanjing}} \\
        \cmidrule(lr){2-5} \cmidrule(lr){6-9} \cmidrule(lr){10-13}
        & \textbf{Precision} & \textbf{Recall} & \textbf{F1-Score} & \textbf{IoU} & \textbf{Precision} & \textbf{Recall} & \textbf{F1-Score} & \textbf{IoU} & \textbf{Precision} & \textbf{Recall} & \textbf{F1-Score} & \textbf{IoU} \\
        \midrule
        Traffic Sign & 0.66 & 0.61 & 0.64 & 0.74 & 0.37 & 0.39 & 0.38 & 0.47 & 0.58 & 0.54 & 0.56 & 0.44 \\
        Street Light & 0.89 & 0.94 & 0.91 & 0.80 & 0.66 & 0.88 & 0.75 & 0.59 & 0.51 & 0.29 & 0.37 & 0.29 \\
        Signal Light & 0.92 & 0.77 & 0.84 & 0.78 & 0.95 & 0.89 & 0.92 & 0.79 & 0.56 & 0.68 & 0.61 & 0.55 \\
        Surveillance Camera & 0.89 & 0.92 & 0.90 & 0.80 & 0.50 & 0.55 & 0.52 & 0.56 & 0.42 & 0.34 & 0.37 & 0.28 \\
        Cylindrical Bollard & 0.78 & 0.91 & 0.84 & 0.67 & 0.49 & 0.57 & 0.53 & 0.53 & 0.73 & 0.69 & 0.71 & 0.57 \\
        Fire Hydrant & 0.86 & 0.75 & 0.80 & 0.63 & 0.81 & 0.81 & 0.81 & 0.68 & 0.33 & 0.25 & 0.29 & 0.39 \\
        Trash Bin & 1.00 & 0.89 & 0.94 & 0.80 & 0.77 & 0.97 & 0.86 & 0.69 & 0.85 & 0.94 & 0.89 & 0.65 \\
        Manhole & 0.62 & 0.19 & 0.30 & 0.21 & 0.34 & 0.52 & 0.41 & 0.36 & 0.26 & 0.20 & 0.22 & 0.21 \\
        Traffic Cone & 0.60 & 0.43 & 0.50 & 0.49 & 0.46 & 0.30 & 0.36 & 0.50 & 0.65 & 0.39 & 0.49 & 0.44 \\
        Spherical Bollard & 0.88 & 0.70 & 0.78 & 0.59 & 0.77 & 0.77 & 0.77 & 0.66 & 0.69 & 0.55 & 0.61 & 0.43 \\
        \midrule
        \textbf{Average} & \textbf{0.81} & \textbf{0.71} & \textbf{0.74} & \textbf{0.65} & \textbf{0.61} & \textbf{0.66} & \textbf{0.63} & \textbf{0.58} & \textbf{0.56} & \textbf{0.49} & \textbf{0.51} & \textbf{0.42} \\
        \bottomrule
        \end{tabular}%
        }
    \end{table*}

    \subsection{3D Point Cloud Segmentation}
        \textbf{Experimental Setup and Baselines.} In this benchmark, we adopt the training-free open-set 3D inventory pipeline proposed by Liu et al. \citep{liu2025training}. The input of this stage is the 2D detection output produced by the FT-G-DINO model in Task 1. Specifically, detected boxes are utilized as geometric prompts for foundation vision models (e.g., SAM) to extract 2D instance masks in panoramic images, which are then projected into 3D point clouds through calibrated image-point cloud alignment and multi-view geometric fusion to obtain category-wise 3D instance masks.

        \textbf{Results and Analysis.} Quantitative results are reported in Table~\ref{tab:pc_seg_compact}. The method achieves the best overall performance in Wuhan (Average F1 = 0.74, mIoU = 0.65), followed by Shanghai (Average F1 = 0.63, mIoU = 0.58), while Nanjing remains the most challenging city (Average F1 = 0.51, mIoU = 0.42). This performance gap is driven not only by inter-city appearance shifts but also by localized data degradations, such as heavy occlusion and motion blur in the Nanjing subset. In particular, blurred imagery reduces the quality of upstream detections, especially for small or distant objects such as surveillance cameras and fire hydrants, which subsequently weakens the image-to-point-cloud projection stage. Moreover, many street lights in Nanjing are heavily occluded by trees. This issue is also reflected in Table~\ref{tab:loc_results_final}, where Street Light exhibits the worst localization accuracy among all categories and cities, with an error reaching 0.53 m. Across classes, stable performance is observed on geometrically distinctive objects such as Traffic Sign, Signal Light, and Trash Bin, whereas ground-level or small-scale targets (e.g., Manhole and Traffic Cone in some cities) show lower recall and F1, suggesting that detection completeness, occlusion handling, and cross-view projection consistency remain the main bottlenecks for methods based on image and point cloud fusion.

        \textbf{Key Findings.} Zero-shot 3D segmentation via multi-modal projection is highly promising, yet its performance is heavily bounded by the quality of upstream 2D detections. Addressing severe occlusions and cross-view projection consistency remains the primary bottleneck for complex urban environments.

    \begin{table*}[t]
        \centering
        \caption{Benchmark results of Attribute Recognition on the WHU-Infra3D dataset. The metric is Attribute Accuracy (\%). \textbf{Avg.} denotes the average accuracy across all categories. Category abbreviations are defined in Section~\ref{subsec:data_annotation}. The best results are highlighted in \textbf{bold}.}
        \label{tab:attribute_benchmark}
        \resizebox{\textwidth}{!}{%
        \begin{tabular}{l:c:cccccccccc}
        \toprule
        \textbf{Method} & \textbf{Avg.} & \textbf{Tra. Sign} & \textbf{Str. Light} & \textbf{Sig. Light} & \textbf{Surv. Cam.} & \textbf{Cyl. Bol.} & \textbf{Hydrant} & \textbf{Trash Bin} & \textbf{Manhole} & \textbf{Tra. Cone} & \textbf{Sph. Bol.} \\
        \midrule
        LLaVA (Zero-shot) & 65.29 & 57.70 & 85.15 & 65.06 & 76.60 & 54.71 & 94.28 & 63.95 & 66.73 & 66.05 & 68.55 \\
        GPT-4o (Zero-shot) & 77.75 & 78.94 & 90.49 & 72.71 & 86.59 & 67.37 & 98.24 & 55.27 & 81.38 & 74.48 & 62.56 \\
        Qwen-VL (Zero-shot) & 78.16 & 78.80 & 92.83 & 71.64 & 84.70 & 63.53 & 99.00 & 52.76 & 82.43 & 79.21 & 71.67 \\
        \midrule
        Qwen-VL (Fine-tuned) & 83.26 & 78.79 & 92.05 & 72.99 & 89.06 & 87.03 & 99.23 & 93.03 & 94.25 & 85.66 & 87.37 \\
        \textbf{Qwen-VL (FT + RAG)} & \textbf{91.30} & \textbf{91.72} & \textbf{98.14} & \textbf{82.68} & \textbf{88.67} & \textbf{92.57} & \textbf{99.63} & \textbf{95.29} & \textbf{97.16} & \textbf{94.01} & \textbf{90.20} \\
        \bottomrule
        \end{tabular}%
        }
    \end{table*}

    \begin{table*}[t]
        \centering
        \caption{Attribute-level accuracy (\%) on selected difficult
        attributes. The best result per row is highlighted in \textbf{bold}.}
        \label{tab:hard_attr}
        \resizebox{\textwidth}{!}{%
        \begin{tabular}{llp{7.8cm}:ccc:cc}
        \toprule
        \textbf{Category} & \textbf{Attribute} & \textbf{Description}
            & \textbf{LLaVA} & \textbf{GPT-4o} & \textbf{Qwen-VL}
            & \textbf{Qwen-VL (FT)} & \textbf{Qwen-VL (FT+RAG)} \\
        \midrule
        Cyl. Bol.   & Reflect   & Whether reflective strips are present \{Yes, No\}
            &  2.36 & 17.65 & 15.06 & 94.40 & \textbf{98.40} \\
        Manhole    & Pattern   & Cast pattern on the cover surface \{Rain, Power, ...\}
            & 10.71 & 17.24 & 43.75 & 95.65 & \textbf{100.00} \\
        Sig. Light & Type      & Applicable traffic participant \{Pedestrian, Motor, ...\}
            & 32.98 & 58.95 & 66.24 & 69.23 & \textbf{79.21} \\
        Tra. Sign   & Type      & Regulatory function of the sign \{Prohibit, Warning, ...\}
            & 30.86 & 59.21 & 81.02 & 70.13 & \textbf{94.33} \\
        Surv. Cam.  & Shape     & Physical form of the camera \{Dome, Box, ...\}
            & 43.75 & 68.75 & 86.40 & 86.21 & \textbf{92.24} \\
        \bottomrule
        \end{tabular}%
        }
    \end{table*}

    \subsection{Attribute Recognition}
        \textbf{Experimental Setup and Baselines.} For this benchmark, 2D target ROIs cropped from panoramic images in the Core Set are utilized as visual inputs. Under an \textbf{In-domain} random split protocol, we evaluate representative Large Vision-Language Models (LVLMs) as generative baselines following the evaluation framework proposed in~\cite{Fu2026}. Specifically, we include \textbf{GPT-4o}~\cite{hurst2024gpt4o}, \textbf{LLaVA}~\cite{liu2024llava}, and \textbf{Qwen-VL}~\cite{bai2023qwenvl} as representative models spanning a range of architectures and training paradigms. These models are evaluated under three settings: Zero-shot Inference, Fine-Tuning (LoRA)~\cite{hu2022lora}, and Retrieval-Augmented Generation (RAG)~\cite{lewis2020rag} (which utilizes the training set as an external knowledge base as detailed in~\cite{Fu2026}), to comprehensively assess their capability in diagnosing statuses of urban infrastructure components.
        
        \textbf{Results and Analysis.} The quantitative results are presented in Table \ref{tab:attribute_benchmark}. Among generative models, general-purpose LVLMs like GPT-4o demonstrate strong zero-shot performance (77.75\%), outperforming LLaVA (65.29\%) due to its stronger multimodal reasoning capability. Meanwhile, Qwen-VL achieves slightly better zero-shot performance (78.16\%), indicating its advantage in visual-language alignment even without task-specific training. After fine-tuning, Qwen-VL further improves to 83.26\%, demonstrating the effectiveness of domain-specific adaptation. Notably, Qwen-VL with RAG achieves the state-of-the-art average accuracy of 91.30\%, confirming that integrating domain-specific adaptation with external knowledge retrieval is essential for professional infrastructure inspection.
        
        Furthermore, Table~\ref{tab:hard_attr} details model performance on a selected set of particularly challenging attributes across multiple object categories. Zero-shot models consistently fail on attributes requiring domain-specific visual priors, such as \textit{Cylindrical Bollard Reflective Attribute} and \textit{Manhole Surface Pattern}, where all three models score below 44\%; domain adaptation via fine-tuning and RAG recovers performance substantially.
        \textit{Signal Light Type} and \textit{Traffic Sign Type} remain challenging throughout, as fine-grained semantic distinctions under varying viewpoints are difficult to resolve even after adaptation. \textit{Surveillance Camera Shape} improves progressively across
        settings yet still leaves room for further gains.
        
        \textbf{Key Findings.} General-purpose LVLMs possess foundational cognitive potential, but accurately diagnosing long-tailed and domain-specific infrastructure attributes strictly requires domain-specific fine-tuning coupled with external knowledge retrieval (RAG).

    \section{Challenges and Future Work}

    \begin{figure}[!t]
        \centering
        \includegraphics[width=\columnwidth]{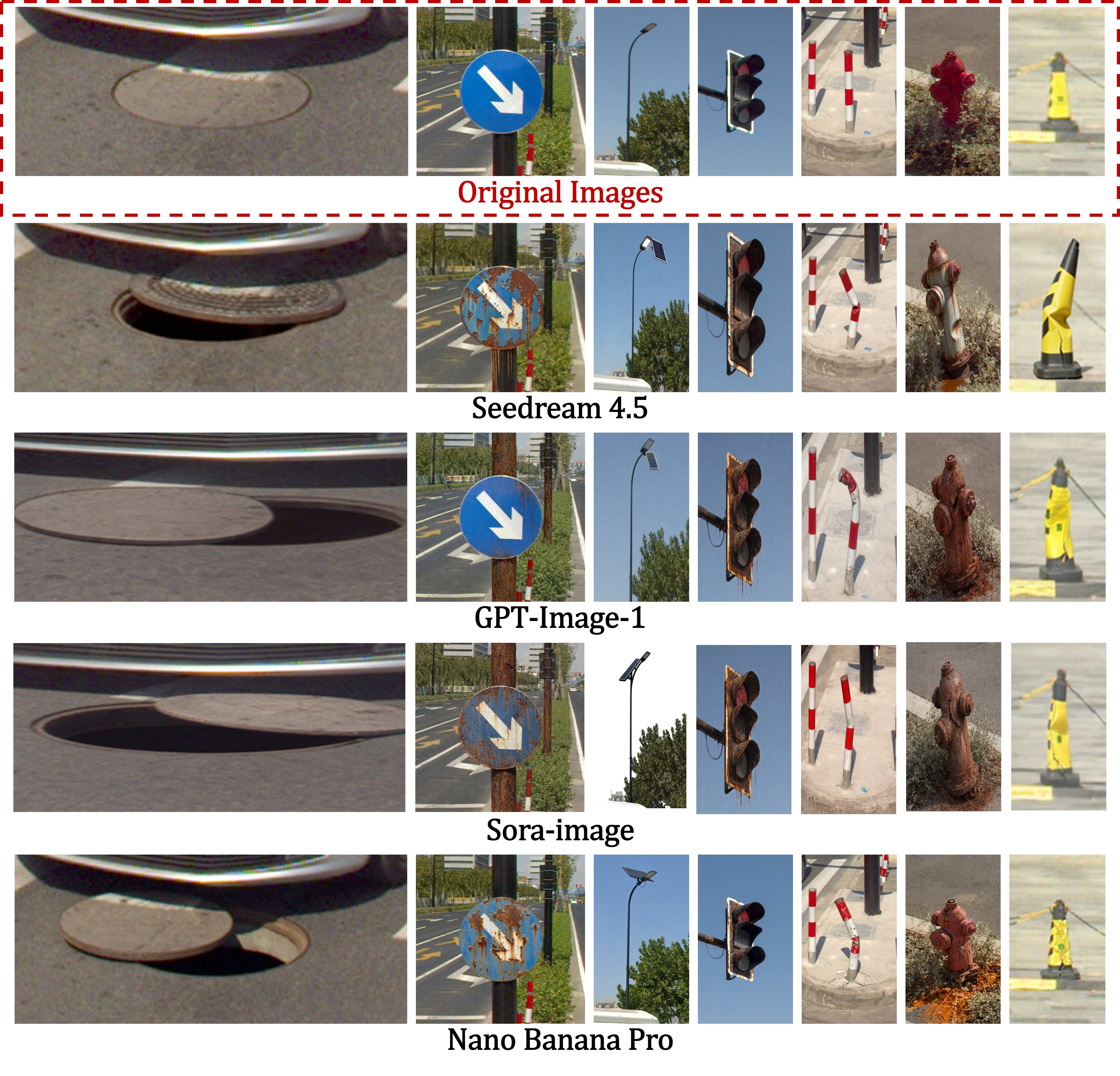}
        \caption{Illustrative examples of rare-status synthesis in infrastructure scenes. The top row shows original infrastructure images, while the subsequent rows present text-guided edited/generated results from Seedream 4.5, GPT-Image-1, Sora-image, and Nano Banana Pro, conditioned on status descriptions such as displaced/open manhole cover, severe rust, and damaged traffic cone.}
        \label{fig:generated_status}
    \end{figure}
   
    \subsection{Long-tailed Status Recognition}
        As highlighted by the poor zero-shot performance on challenging cases in Table~\ref{tab:hard_attr}, many critical infrastructure statuses---such as specific material patterns, severe physical damage, or rust---inherently follow a long-tailed distribution. This natural scarcity of positive samples poses a significant challenge for data-driven perception, as models often fail to generalize these rare concepts or easily overfit the few available examples during training. To mitigate this fundamental data imbalance, a promising future direction is to leverage recent advancements in generative AI. As illustrated in Figure~\ref{fig:generated_status}, real infrastructure crops can be used as references for text-guided status editing/generation with foundation models to synthesize diverse hard cases under rare defect statuses. In future work, this strategy could enrich training data for better long-tail robustness and support the construction of controlled stress-test sets for evaluating the reliability of attribute diagnosis algorithms.

    \begin{figure*}[t]
        \centering
        \includegraphics[width=\textwidth]{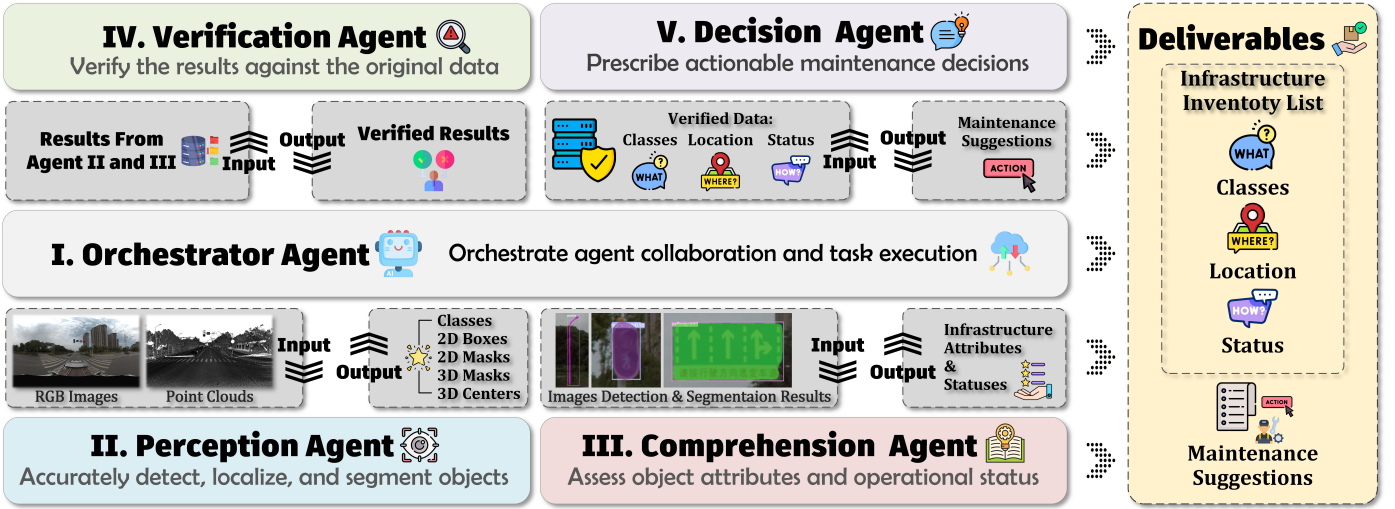}
        \caption{A conceptual framework for autonomous infrastructure asset management driven by Multi-Agent collaboration. The workflow transitions from passive perception to active decision-making under the coordination of a central orchestrator and four specialized agents. WHU-Infra3D provides the multi-modal training and evaluation substrate for the perception, comprehension, and verification modules.}
        \label{fig:agent_framework}
    \end{figure*}

    \subsection{Towards Autonomous Asset Management Agents}
        Current infrastructure inventory pipelines are inherently linear, relying on isolated perceptual modules while heavily depending on human intervention for verification and maintenance planning. To bridge this gap, a promising frontier is the development of autonomous Multi-Agent Systems (MAS), as conceptualized in Figure~\ref{fig:agent_framework}. Under this paradigm, the static inventory workflow is reimagined as an interactive, closed-loop ecosystem orchestrated by a central \textbf{Orchestrator Agent}, which delegates tasks across four specialized sub-agents.
        
        Within this ecosystem, a \textbf{Perception Agent} executes geometry- and instance-centric primitives (e.g., detection, cross-view matching, 3D localization, and point cloud segmentation), while a \textbf{Comprehension Agent} is allocated for high-level semantic reasoning, such as attribute parsing and defect diagnosis. Crucially, to mitigate the hallucination risks inherent in foundation models, a VLM-driven \textbf{Verification Agent} acts as a digital quality inspector. Rather than passively passing data forward, it establishes a dynamic feedback loop: by cross-checking visual evidence against 3D geometric topologies, it can autonomously prompt upstream agents to re-evaluate ambiguous targets when inconsistencies (e.g., false positives or logic conflicts) arise. Ultimately, the validated structural digital twin is handed to a \textbf{Decision Agent}, which synthesizes Retrieval-Augmented Generation (RAG) with civil engineering codes to formulate actionable operations, such as maintenance prioritization and repair strategies.
        
        Under this visionary perspective, the primary objective of WHU-Infra3D is not to deliver a turnkey MAS pipeline, but to serve as the indispensable data catalyst. In essence, the diverse benchmark tasks formulated in Section 4 mirror the foundational capabilities demanded by the Perception and Comprehension agents. By concurrently anchoring 2D/3D sensing, cross-modal association, and visual-linguistic reasoning, the dataset provides a rigorous, unified testbed to incubate and evaluate future autonomous infrastructure management workflows.
        
    \subsection{Multi-modal Data Synergy}
        A practical bottleneck in current infrastructure inventory pipelines is that image-point cloud fusion remains weak in the perception stage. As exemplified by recent training-free workflows \citep{liu2025training}, the dominant paradigm still performs detection and segmentation primarily in image space and then projects results into 3D point clouds, rather than jointly reasoning over both modalities from the outset. This design is effective for rapid deployment, but it underuses point-cloud-native cues in downstream diagnosis. In particular, current attribute and status recognition is still largely image-dependent, while some engineering-relevant attributes are strongly correlated with LiDAR measurements: for example, whether a bollard has reflective material is closely related to intensity patterns, and geometric inconsistencies in local 3D structure can provide direct evidence for physical damage. Therefore, a key direction is not merely adding more labels, but building tighter image-point cloud coupling for both perception and status assessment.

    \section{Conclusion}
        In this paper, we introduced \textbf{WHU-Infra3D}, a large-scale multi-modal benchmark tailored to advance roadside infrastructure inventory from perceptual mapping to fine-grained cognitive management. By thoroughly pairing high-resolution panoramic images with dense LiDAR point clouds via 2D-3D instance association and global cross-frame tracking, it offers a geometrically and semantically consistent data foundation. Crucially, its inclusion of over 181k detailed structural and status annotations effectively addresses the data scarcity that has long hindered automated status diagnosis.

        We conducted comprehensive evaluations across five interconnected tasks: 2D detection, 2D cross-view matching, 3D geo-identification, open-set 3D segmentation, and multi-modal attribute reasoning. Our results reveal that while current models achieve promising in-domain diagnostic metrics, they severely lack robustness against unseen domain shifts and struggle to resolve long-tailed defective instances. This underscores that modern pipelines still underutilize the synergistic potential of image-point cloud interactions for comprehensive scene interpretation. By systematically exposing these bottlenecks, WHU-Infra3D bridges a critical gap in infrastructure-focused computer vision and serves as a vital testbed for future innovations, including robust open-vocabulary generalization, generative data synthesis, and autonomous agent-driven management workflows.

    \section *{Declaration of Interests}
        The authors declare that they have no known competing financial interests or personal relationships that could have appeared to influence the work reported in this paper.
    
    \section *{Acknowledgment}
        This work was supported by the National Natural Science Foundation of China (No. 42571521) and the Key R\&D Program of Hubei Province (No. 2025BEB011).
        
\bibliography{rmfile}

\end{document}